\documentclass[11pt,x11names]{article}
\usepackage{setspace}
\usepackage[final]{misc/acl}

\usepackage{times}
\usepackage{latexsym}

\usepackage[T1]{fontenc}
\usepackage[utf8]{inputenc}

\usepackage{microtype}

\usepackage{inconsolata}

\usepackage{graphicx}

\usepackage{xcolor}
\usepackage{subcaption}  % Required for subfigures
\usepackage{booktabs}
\usepackage{todonotes}
\usepackage{amsmath,amssymb}
\usepackage[capitalize]{cleveref}
\usepackage{xspace}
\usepackage{soul} % for \ul and \TODOMARK
\usepackage{caption}
\usepackage{multirow}
\usepackage{ulem}
\usepackage{float}
\usepackage{array}
\usepackage{bm}
\usepackage{xfrac}
\usepackage{tcolorbox} % for frames
\usepackage[shortlabels]{enumitem}
\usepackage[outline]{contour}% http://ctan.org/pkg/contour
\usepackage{tikz}
\usepackage{float}

\setlist[itemize]{noitemsep,left=0mm}
\usepackage{stfloats} 
\usepackage{tabularx}

\usepackage{marginnote}

\makeatletter
\renewcommand{\paragraph}{%
  \@startsection{paragraph}{4}{\z@}%
    {0pt} % Space above
    {-\baselineskip} % Space below
    {\normalfont\normalsize\bfseries}%
}
\makeatother

\usepackage{adjustbox}

\usepackage{enumitem}
\makeatletter\def\Hy@Warning#1{}\makeatother
\let\svthefootnote\thefootnote
\newcommand\blankfootnote[1]{%
  \let\thefootnote\relax\footnotetext{#1}%
  \let\thefootnote\svthefootnote%
}

\graphicspath{{img/}}
\title{
   Contrastive Learning for Task-Independent SpeechLLM-Pretraining
}

\author{ Maike Züfle \phantom{\and} Jan Niehues \\
        Karlsruhe Institute of Technology, Germany \\
        \texttt{\{maike.zuefle, jan.niehues\}@kit.edu}
        }

\begin{document}

\twocolumn

\maketitle

\maketitle

\begin{abstract}
Large language models (LLMs) excel in natural language processing but adapting these LLMs to speech processing tasks efficiently is not straightforward. Direct task-specific fine-tuning is limited by overfitting risks, data requirements, and computational costs. To address these challenges, we propose a scalable, two-stage training approach: (1) A task-independent speech pretraining stage using contrastive learning to align text and speech representations over all layers, followed by (2) a task-specific fine-tuning stage requiring minimal data. This approach outperforms traditional ASR pretraining and enables the model to surpass models specialized on speech translation and question answering while being trained on only 10\% of the task-specific data.
\end{abstract}

\blankfootnote{\url{https://github.com/MaikeZuefle/contr-pretraining}}

\section{Introduction}
\label{sec:introduction}

Large language models (LLMs) have demonstrated impressive capabilities in language understanding and translation tasks \citep{openai2024gpt4technicalreport, dubey2024llama3herdmodels}. However, their reliance on written text limits their application in real-world scenarios, where communication often occurs in the form of speech. Extending LLMs to spoken language could enable them to reason about and process spoken content effectively.

A straightforward approach is to use a cascaded pipeline \citep{shen2023hugginggpt} where speech is first converted to text using an automatic speech recognition (ASR) model, and the text is subsequently processed by an LLM. While simple, this approach has limitations: it cannot be trained end-to-end, ASR errors can propagate through the pipeline, and critical paralinguistic information, such as speaker features or noise, is lost during transcription.

To address these issues, recent works have proposed SpeechLLMs \citep{wang2024blspbootstrappinglanguagespeechpretraining, tang2024salmonn, nguyen2024spiritlminterleavedspoken, chu2024qwen2audiotechnicalreport}, which connect speech encoders to LLMs via a trainable projector in an end-to-end manner. However, training them effectively remains a key challenge due to the scarcity of labelled data for downstream tasks \citep{xu-recent-advances-st, unlu-menevse-etal-2024-dealing}.

Pretraining SpeechLLMs before finetuning on task-specific data can help overcome this challenge, reducing the need for large downstream datasets. A straightforward strategy is to use ASR data, training on the ASR task itself \citep{tang2024salmonn, wang2024freezeomnismartlowlatency}.
% or aligning next-word prediction (NWP) behaviours for parallel speech and text inputs \citep{wang2024blspbootstrappinglanguagespeechpretraining}. 
However, this might lead to overfitting to this pretraining task \citep{tang2024salmonn}. 

\begin{figure*}[!t]
\centering
\includegraphics[width=1.0\linewidth]{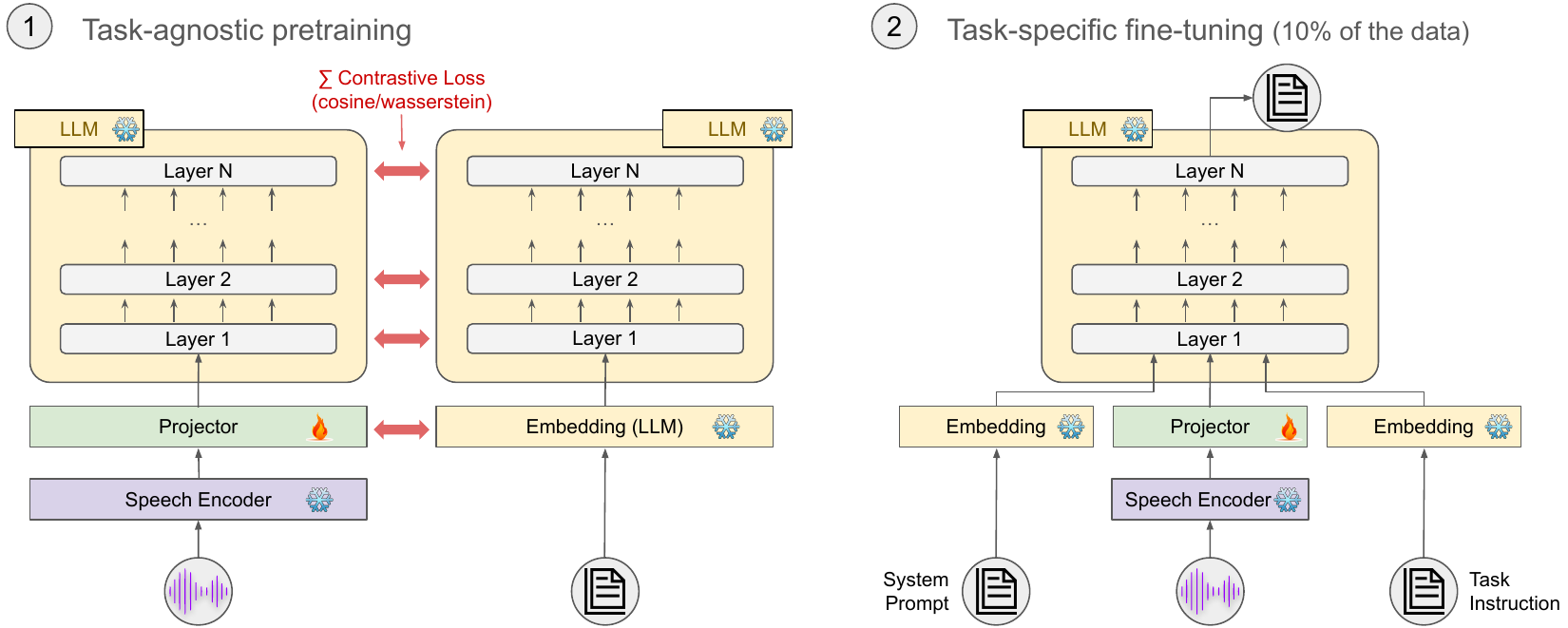}
\caption{An overview of our two-stage training process. First we perform task-agnostic pretraining, by leveraging ASR data to calculate a contrastive loss over a selected set of layers. Then we finetune on task-specific data. The speech encoder and the LLM remain frozen (indicated by the snowflake), we only train the projector. }
\label{fig:highlevel_schema}
%\vspace{-0.3cm}
\end{figure*}

Our work introduces contrastive learning as a task-independent pretraining strategy for SpeechLLMs using ASR data. Contrastive learning aligns the representations of paired speech and text inputs while encouraging representations of unrelated inputs to remain distinct. We demonstrate that applying a contrastive loss to the representations after every layer and pretraining on just 400 hours of parallel speech-text data serves as an effective, task-agnostic foundation for SpeechLLMs.

To evaluate our approach, we finetune the pretrained model on three downstream tasks: ASR, speech translation (ST), and speech question answering (SQA). We compare its performance against that of traditional ASR pretraining and next-word prediction (NWP) pretraining using mixed speech-text input. The contrastive strategy not only surpasses the other methods but also matches the performance of task-specific state-of-the-art (SOTA) models on nearly all metrics.  We also analyse combinations of the pretraining methods to explore potential complementary interactions.

Scaling up contrastive pretraining to 1,400 hours 
yields further performance gains, with our models surpassing specialized models and current SpeechLLMs. Even in simulated low-resource scenarios, 
using only 10\% of the finetuning data,
our method consistently outperforms existing SpeechLLMs.

Our SpeechLLM architecture is simple yet flexible, consisting of a speech encoder, a projector, and an LLM. Only the projector is trained, making the model parameter-efficient and enabling maximum flexibility in adapting to the ever-evolving landscape of LLMs and speech encoders. Moreover, this approach preserves the LLM's text-to-text capabilities, ensuring its functionality across both speech and text-based tasks. An overview of our approach is provided in \cref{fig:highlevel_schema}.

The main contributions of our paper are:
\begin{enumerate}[topsep=1pt,itemsep=-1ex,partopsep=1ex,parsep=1ex]
    \item We propose contrastive pretraining for SpeechLLMs, achieving state-of-the-art results in ST and SQA.
    \item We demonstrate the effectiveness of our pretraining method in low-resource settings.
    \item We provide a comprehensive analysis of contrastive learning for SpeechLLMs and its interaction with other pretraining methods.
    \item We show that contrastive pretraining preserves a model's ability to recognise paralinguistic speech features.
\end{enumerate}

\section{Related Works}\label{sec:background}

\paragraph{SpeechLLMs.}
SpeechLLM architectures typically consist of three main components: a speech encoder, a pretrained language model, and a projector linking the two. Training strategies differ based on which components are finetuned.

\citet{chu2024qwen2audiotechnicalreport} adopt a comprehensive strategy by finetuning all components. Similarly, other works explore finetuning both the projector and the encoder, alongside applying Low-Rank Adaptation (LoRA; \citealt{hu2022lora}) to the LLM \citep{wu2023decoderonlyarchitecturespeechtotextlarge, gong2024listen, fathullah2023promptinglargelanguagemodels}. In contrast, some methods freeze the LLM entirely, focusing on training only the encoder and projector \citep{wang2024freezeomnismartlowlatency, held2024distillingendtoendvoiceassistant, fathullah-etal-2024-audiochatllama}.
Several other approaches focus on finetuning the projector while introducing LoRA layers to the LLM, leaving the speech encoder and LLM frozen \citep{peng2024voicetextblenderaugmentinglargelanguage, tang2024salmonn}. 
An even more lightweight approach freezes both the encoder and the LLM, finetuning only the projection layers \citep{wang2024blspbootstrappinglanguagespeechpretraining}.

In this work, we use the latter approach, using a frozen speech encoder and LLM while finetuning only the projection layers. This preserves the text-to-text abilities of the LLM, while also being parameter efficient. 
% However, our training strategy and text-speech alignment process is different to BLSP, as detailed in the next section.

\paragraph{Text and Speech Alignment.}
While `alignment' has various meanings \citep{hammerl-etal-2024-understanding}, in this work, we use it to refer to the process of creating meaningfully similar representations for text and speech.
One approach to achieve this alignment is to directly finetune on the desired tasks without explicitly modelling alignment \citep{fang-etal-2024-llama-omni, fathullah2023promptinglargelanguagemodels}. An alternative is to use a downstream task to pretrain the SpeechLLM before finetuning on desired tasks, for example pretraining on ASR \citep{tang2024salmonn, wang2024freezeomnismartlowlatency} or ASR variants
\citep{wang2024blspbootstrappinglanguagespeechpretraining}.

Other works achieve alignment by encouraging paired speech and text embeddings closer together using an appropriate loss term. For example, \citet{held2024distillingendtoendvoiceassistant} use an L2 loss on speech and text embeddings in addition to a KL-divergence loss on the output distributions, and \citet{chuang-etal-2020-worse} 
use a cosine distance loss between paired embeddings.
Optimal transport has also been used to align paired embeddings in encoder-decoder speech translation systems \citep{le2023pretrainingspeechtranslationctc, tsiamas-etal-2024-pushing}.

Contrastive training is another widely used approach, bringing parallel text and speech representations closer while pushing non-parallel representations apart. It has been used in knowledge distillation for spoken language understanding \citep{zhu22f_interspeech, cappellazzo-etal-2024-continual}, as an auxiliary loss for encoder-decoder ST systems \citep{ye-etal-2022-cross, ouyang-etal-2023-waco},
and for music \citep{manco2022}. It has also been applied to classification tasks, such as emotion recognition \citep{sachidananda2022calmcontrastivealignedaudiolanguage, zhou-etal-2024-clasp}. 

%alignment using mixed input
Other approaches mix text and speech in the input to help models learn both modalities. For example, \citet{fang-etal-2022-stemm} align the output distributions of speech and mixed text-speech inputs, while \citet{nguyen2024spiritlminterleavedspoken} and \citet{peng2024voicetextblenderaugmentinglargelanguage} mix speech and text sequences for next-word prediction and question-answering tasks.

In this work, we leverage the previous success of contrastive learning on speech models and apply it to SpeechLLMs. We show how contrastive learning can be applied successfully in this new scenario and combined with other losses. In contrast to other works, we experiment with using optimal transport for the contrastive loss function and apply the loss to several hidden layers of the SpeechLLM.

\section{Methods}\label{sec:Methods}
We propose contrastive learning as a task-independent pretraining method for aligning speech and text representations, leveraging abundantly available ASR data. We compare with two ASR-based pretraining baselines: traditional ASR pretraining and next-word prediction with mixed text-speech spans, which we also introduce as a pretraining approach in this paper.

\paragraph{Notation.}Let $\mathbf{t} \in \mathbb{R}^{H \times N}$ represent a text sequence with $N$ tokens, where each token is embedded into a $H$-dimensional space. The speech sequence is embedded into $M$ vectors, with $\mathbf{s} \in \mathbb{R}^{H \times M}$. Note that usually $M > N$. 

\subsection{Contrastive Training}\label{subsec:method_contr}

Contrastive training brings representations of matched speech-text pairs closer whilst pushing representations of mismatched pairs apart.
Harnessing both matched and mismatched pairs is often a very efficient way of using a dataset and allows one to leverage abundantly available ASR data.

In this paper, we employ the InfoNCE loss \citep{oord2019representationlearningcontrastivepredictive} similar to other works \citep{manco2022}.
InfoNCE loss is defined as:
\begin{equation}
\mathcal{L}_{\text{InfoNCE}} = -\frac{1}{|B|} \sum_{i \in B} \log \frac{\exp\left( \frac{\text{sim}(\mathbf{s}_i, \mathbf{t}_i)}{\tau} \right)}{\sum_{j \in B} \exp\left( \frac{\text{sim}(\mathbf{s}_i, \mathbf{t}_j)}{\tau} \right)},
\end{equation}
where $\text{sim}(\cdot, \cdot)$ denotes a similarity measure, $\mathbf{s}_i$ and $\mathbf{t}_i$ denote speech and text embeddings respectively for the $i$-th example in the batch, and  $\tau$ is the temperature parameter controlling the sharpness of the similarity scores. Summing over $B$ in the denominator includes the aligned speech-text pair $ (\mathbf{s}_i, \mathbf{t}_i) $, as well as negative (contrastive) pairs. We choose to use all contrastive pairs within a mini-batch to allow for an efficient implementation.

The contrastive loss can also be calculated for multiple layers throughout the network and then summed. \cref{fig:highlevel_schema} illustrates this setup.
We find that this gives the best results (see \cref{sec:results}). 

We adopt two different similarity measures:
\paragraph{Cosine Similarity.}
One way to compute $\text{sim}(\mathbf{s}_i, \mathbf{t}_i)$ is to use cosine similarity. Since the length of the speech sequences, $M$, is typically much larger than the length of the text sequences, $N$, we cannot directly compute cosine similarity on the entire sequence. Instead, we average the embeddings over the sequence length to obtain a single embedding for each
\citep{ye-etal-2022-cross}.
We denote this pretraining strategy by \textit{contr-cos}.
\paragraph{Wasserstein Distance.}
To avoid 
aggregating the sequences, we can leverage the Wasserstein (or Optimal Transport) distance between two sequences \citep{peyre-ot-2019, le2023pretrainingspeechtranslationctc}.  In our case, the Wasserstein distance involves finding the optimal way to "transport" mass from one normalized set of embeddings to another, enabling us to compute the alignment between text and speech embeddings of differing lengths.

More precisely, the optimal transport distance is calculated as follows: a mass of $1/N$ is placed at each text embedding location (points in a $H$-dimensional space) and we want to move this total mass of $1$ to the $M$ speech embedding locations, $1/M$ to each. There is a cost $C_{ij}$ associated with moving a unit of mass from location $t_i$ to location $s_j$ for all $i \in \{1, \ldots, N\}$ and $j \in \{1, \ldots, M\}$. The cost is given by the $L^p$ distance in the embedding space. This can be written as:
\begin{align}
\label{eq:wasserstein_seq}
d_{\text{Wasser}}(\textbf{s}, \textbf{t}) = \min_{\textbf{Z}} \{ \textbf{C}(\textbf{s}, \textbf{t}) \odot \textbf{Z} \}, \text{ s.t. } \nonumber \\ 
\textbf{Z} \mathbf{1}^N = \mathbf{1}^N/N, \textbf{Z}^T \mathbf{1}^M = \mathbf{1}^M/M, \textbf{Z} \geq \textbf{0},
\end{align}
where $\textbf{Z}$ is the transportation matrix with row and column sums $\frac{1}{N}$ and $\frac{1}{M}$ respectively, $\textbf{C}(\textbf{s}, \textbf{t})$ is the cost matrix for sequences $\textbf{s}$ and $\textbf{t}$, $\mathbf{1}^N$ denotes a vector of $1$'s, and $\odot$ denotes the entrywise dot product.
For a more detailed description, we refer the reader to \citet{le2023pretrainingspeechtranslationctc}.

Using the Wasserstein distance in the loss is expensive, so we resort to an upper-bound approximation that can be computed efficiently using the Sinkhorn algorithm \citep{Sinkhorn1967ConcerningNM}. Note, however, that even with this approximation, using Wasserstein distance is more computationally expensive ( $O(n^2/epsilon^2)$ \citep{pham2020unbalancedoptimaltransportanalysis})  than using cosine similarity ($O(n)$).

We use the negative distance as the similarity measure for $\mathcal{L}_{\text{InfoNCE}}$ and  denote this pretraining strategy by \textit{contr-wasser}.

\subsection{Other Pretraining Approaches}
To evaluate our contrastive pretraining approach we compare it with traditional ASR pretraining and a mixed text-speech NWP pretraining approach.
\paragraph{ASR Pretraining.} We use task-specific fine-tuning on the ASR task, i.e. following the ASR-instruction prompt, for pretraining  \citep{tang2024salmonn, wang2024freezeomnismartlowlatency} and use it as a baseline.

\paragraph{Mixed Next Word Prediction.}\label{subsec:mixed_nwp}
Inspired by \citet{nguyen2024spiritlminterleavedspoken}, 
who add mixed NWP to their finetuning data mix for downstream tasks,
we propose mixed text and speech NWP as a pretraining strategy.
We mix text and speech embeddings across randomly selected spans in the sequence, maintaining coherence with word boundaries. A sample sequence might look as follows:  $t_0^0, t_1^0, s_2^1, s_3^1, s_4^2, s_5^2, s_6^2,  t_7^3$, where superscripts represent word boundaries, and subscripts represent subwords within the text or speech embedding. Since speech embeddings typically differ in length from text embeddings, we compute the NWP loss exclusively on text tokens.
Details on how we select the speech and text spans are provided in \cref{app:mixed_nwp}.

We also experiment with using these mixed sequences for contrastive learning: 
instead of aligning speech and text, we align mixed text-speech representations with text representations.

\section{Experiments}\label{sec:experiments}

Our experiments follow a two-stage training strategy. First, we perform text-speech alignment pretraining using ASR data to adapt the model to handle speech. In the second stage, we finetune the model jointly on a mix of task-specific data. This approach leverages abundant ASR data to reduce reliance on large task-specific datasets, enabling effective adaptation even in low-resource settings.

\subsection{Model Architecture}\label{subsec:exp_model} Our SpeechLLM architecture consists of three core components: a speech encoder, a text-based LLM, and a projector bridging the two. 
We select HuBERT \citep{hubert-2021} as our speech encoder due to its ability to process audio inputs longer than 30 seconds, which is crucial for SQA tasks. 
Recent research \citep{hassid2024textuallypretrainedspeechlanguage, nguyen2024spiritlminterleavedspoken} has demonstrated strong performance with HuBERT variants. We specifically utilize the \texttt{facebook/hubert-large-ls960-ft} variant.
We use \texttt{Llama-3.1-8B-Instruct} \citep{dubey2024llama3herdmodels} as our LLM.
As a projector, we choose Q-Former \citep{Li2023BLIP2BL}, a window-level query Transformer, which already proved successful for other SpeechLLMs \citep{tang2024salmonn, held2024distillingendtoendvoiceassistant}. 

For contrastive learning, we compare speech and text embeddings without the use of system prompts. This creates a positional mismatch between contrastive pretraining and finetuning. To address this potential overfitting problem, we experiment with varying the absolute starting position in the positional embeddings. However, this does not consistently improve model performance (\cref{app:layers}). 

\subsection{Training Parameters} In both pretraining and task-specific finetuning, we keep the LLM and speech encoder frozen, adjusting only the speech projector, which has 42.5 million parameters. Training is conducted on four NVIDIA A100-SXM4-40GB GPUs.
ASR pretraining takes approximately 9h, contrastive pretraining with cosine similarity over all layers takes around 5h. The task-specific finetuning takes around 30h with 100\% of the data. 
Detailed training parameters and model specifics are listed in \cref{app:hyper}.

\subsection{Data} 
\paragraph{Pretraining.} For pretraining, we use the English portion of the widely-used MustC-v1 \citep{di-gangi-etal-2019-must} data (approximately 400 hours) to demonstrate effective performance with limited data. MustC includes TED talks with diverse speakers, accents, and topics, along with manually curated transcriptions. Additionally, we test pretraining on a larger dataset, adding the M-version of Gigaspeech \citep{GigaSpeech2021} to our pretraining data, which includes 1,000 hours of speech from audiobooks, podcasts, and YouTube videos.
\paragraph{Finetuning.} We jointly finetune and evaluate our contrastive pretraining method on three diverse tasks covering a range of speech-text model abilities: Automatic Speech Recognition (ASR), Speech Translation (ST), and Speech Question Answering (SQA). For ASR and ST, we use the MustC-v1 dataset for training and testing.  As before, English transcriptions serve as the ASR data, and we use language pairs en-de, en-fr, en-it, and en-es for ST. For question answering, we use Spoken-SQuAD \citep{lee2018spoken}, an English dataset based on Wikipedia, derived from the Stanford Question Answering Dataset \citep[SQuAD]{rajpurkar2016squad100000questionsmachine}.  

While there is overlap between pretraining and finetuning data for the ASR and ST tasks, we consider this a realistic scenario: In practice, when task-specific data is limited, leveraging all available data—even if it appears in both stages—is often necessary. This is particularly relevant for speech translation, where transcripts are abundant but parallel translations are much scarcer.
\paragraph{Paralinguistic Feature Analysis.}
% When finetuning with only 10\% of the data, we randomly sample a subset of each training set.
We also analyse whether contrastive pretraining harms a model's ability to capture paralinguistic features, such as speaking pace, gender or noise. We use two datasets: the \textit{mls-eng-speaker-descriptions} dataset \citep{Pratap2020MLSAL, lacombe-etal-2024-dataspeech, lyth2024natural}, which involves reasoning about speaking rate, gender, background noise, and other features, and the MuST-SHE \citep{bentivogli-etal-2020-gender} dataset, where gender information is needed to correctly translate the segment.

More information about the train and test sets can be found in \cref{app:data} and details on prompts in \cref{app:prompts}.

\subsection{Evaluation}\label{subsec:exp_eval}
We report the following evaluation metrics: WER (Word Error Rate, using \texttt{jiwer}) for ASR, sacreBLEU \citep{post-2018-call} and COMET$^{\text{DA}}_{\text{22}}$ \citep{rei-etal-2020-comet} for ST, and exact match accuracy (EM) and F1 score for SQA. Scores for ST are averaged over the language pairs. For the best performing models, we report results for individual language pairs in \cref{app:more_giga}. 
Moreover, we calculate the average across the three tasks using WER, COMET, and F1 scores. 
Since these scores have different ranges, care has to be taken when averaging. 
We therefore normalize the scores using lower (lb) and upper (ub) bounds as detailed in \cref{subsec:baselines}. We then calculate the normalized average as follows:
\begin{equation}\label{eq:norm_average}
\text{Norm. Avg.} = \frac{1}{|T|} \sum_{t \in T} \frac{\text{score(t)} - \text{lb(t)}}{\text{ub(t)} - \text{lb(t)}},
\end{equation}
where $T$ is the set of tasks, and $score(t)$ is the model's score on task $t$.

We also report the contrastive loss on the test set for the embedding layer in order to evaluate the effectiveness of aligning speech and text. 

\begin{table*}[t]
    \centering

    \resizebox{\linewidth}{!}{%
    \begin{tabular}{clcccccccc}
        \toprule
        \textbf{FT} & \multirow{2}{*}{\textbf{Model}} & \multicolumn{1}{c}{\textbf{ASR}} & \multicolumn{2}{c}{\textbf{ST}} & \multicolumn{2}{c}{\textbf{SQA}}  & \multicolumn{2}{c}{\textbf{Contr loss on test}} & \textbf{Norm.} \\
        
        \cmidrule(lr){3-3} \cmidrule(lr){4-5} \cmidrule(lr){6-7} \cmidrule(lr){8-9}
        
       \textbf{Data} & & \textbf{WER$\downarrow$*} & \textbf{BLEU$\uparrow$} & \textbf{COMET$\uparrow$*} & \textbf{EM$\uparrow$} & \textbf{F1$\uparrow$*}  & \textbf{cos.} & \textbf{wasser.} & \textbf{avg.$\uparrow$}\\
        
        \midrule

        N/A & Specialized &   \hphantom{0}6.54 & 30.99 & 80.02 & 64.19 & 77.10 & N/A & N/A & \hphantom{-}100\\
        N/A  & HuBERT + Llama & 18.38	  & 19.85	 & 73.92	& 36.15  &	54.76 & N/A &N/A &  \hphantom{-}0 \\
        100\% & no pretrain    &12.09 &	28.84	& 79.94	& 64.13	& 76.53	& 1.37	& 1.26	& \hphantom{-}83.08 \\
        \midrule
        \multirow{6}{*}{10\%} & no pretrain   & 23.78	&19.87	&69.72	& 34.72 & 	46.07	& 1.38	& 1.36	& -51.13 \\
         & ASR pretrain  & \textbf{12.21}	&24.82 &	75.70	& 49.48	& 63.36	& 1.37	& 1.30	& \hphantom{-}39.93  \\
         &  contr-cos-emb & 13.98	& 25.47	& 76.32	& 54.86 & 68.49 &	0.97	&0.72 &	\hphantom{-}46.00\\
         & contr-wasser-emb & 15.99	& 26.10	& 77.09	& 54.64 & 67.83 &	1.08 &	0.62 &	\hphantom{-}43.52\\
         & contr-cos-all & 13.06	&27.19	& 78.29	& 60.48	& 73.90	& 1.08	& 0.91	& \hphantom{-}67.39\\
         & contr-wasser-all & 12.92	& \textbf{29.07}	& \textbf{80.52}	& \textbf{64.06}	& \textbf{77.22}	& 1.07	& 0.64	 & \hphantom{-}\textbf{84.96}\\
        \bottomrule
    \end{tabular}%
    }
    \caption{Results of models with no pretraining, ASR pretraining, and contrastive pretraining using cosine similarity (\textit{contr-cos}) or Wasserstein distance (\textit{contr-wasser}) on embedding (\textit{-emb}) or all (\textit{-all}) layers, followed by fine-tuning on a 10\% subset of task-specific data.  The specialized baselines are  Whisper \citep{radford2022robustspeechrecognitionlargescale} for ASR, Seamless \citep{communication2023seamlessm4tmassivelymultilingual} for ST, and \citet{you-etal-2022-end} for SQA. Metrics with * contribute to the normalized average as in \cref{eq:norm_average}.}
    \label{tab:contrastive}  
\end{table*}

\subsection{Baselines}\label{subsec:baselines}

We compare our approach against four baselines: 
(1) Firstly, we take three SOTA models, each specialized on one of the three tasks to set a goal for our models.
Specifically, we use Whisper \citep{radford2022robustspeechrecognitionlargescale} for ASR, Seamless \citep{communication2023seamlessm4tmassivelymultilingual} for ST, and \citet{you-etal-2022-end} for SQA\footnote{More recently, \citet{manakul2024enhancinglowresourcelanguageinstruction} report SQA results for Gemini-1.5-Pro, which underperforms \citet{you-etal-2022-end}.}. These SOTA results  also serve as an upper bound for normalizing the respective scores when calculating the overall (normalized average) performance scores.
(2) As the second baseline, we use the cascaded HuBERT-Llama-3.1-8B-Instruct model, which also serves as the lower bound for calculating overall performance scores.

Our primary objective is to assess the impact of contrastive pretraining. To do so, we need a controlled environment, where we keep the model architecture, data, and training setup as consistent as possible. 
To this end, we include two additional baselines using our training setup: (3) No pretraining and (4) Standard ASR pretraining.

\section{Results}\label{sec:results}

The goal of our experiments is to find the best setup for using contrastive learning for SpeechLLMs. We investigate various contrastive pretraining losses and where to apply them, combinations with other pretraining losses, and the impact of additional data. 
We then compare our best-performing model to recent SpeechLLMs.
Finally, we investigate how text-speech alignment affects the detection of paralinguistic features.

\subsection{Contrastive Pretraining}\label{subsec:contr_pretraining}

\paragraph{Contrastive Loss on the Embedding Layer.} 
We start by calculating the contrastive loss on the embedding layer and comparing the two loss functions: Wasserstein (\textit{contr-wasser-emb}), and cosine similarity (\textit{contr-cos-emb}).
After pretraining, the models are finetuned on three downstream tasks (jointly) using a random 10\% subset of the data, simulating a low-resource setting (\cref{tab:contrastive}).

Our contrastive pretraining models consistently outperform the cascaded Hubert+Llama and no-pretraining baselines across all metrics. In ST and SQA, they also surpass standard ASR pretraining, by over 5\% in SQA. However, ASR-pretrained models remain better for ASR, which is unsurprising, given they are pretrained with more data on this task. Analysing the contrastive loss on the test set, we find that it is significantly lower after contrastive pretraining than after ASR pretraining or direct finetuning on the task, showing that indeed there is more alignment. Interestingly, while contrastive models exhibit lower test-set contrastive loss on the embedding layer, the losses for no-pretraining and ASR-pretraining are similar despite a large performance gap.

\paragraph{Contrastive Loss on Different Layers.} Keeping with the low-resource setting, we investigate calculating the contrastive loss at different LLM layers. We find that applying contrastive loss to deeper layers significantly increases downstream performance, however, the best layer to use is not consistent. 
Therefore, we also explore using a summed loss over all (multiples of 5\footnote{Initial experiments found that using every 5th layer efficiently approximates using all layers  (\cref{tab:multiple_of_5_vs_all}).}) layers in \cref{tab:contrastive}.   
The methods \textit{contr-cos-all} and \textit{contr-wasser-all} achieve normalized task averages of $67.39$ and $84.96$, respectively, significantly outperforming the ASR-pretraining baseline and the \textit{contr-emb} models. \textit{Contr-wasser-all}, despite using only 10\% of the finetuning data, 
even matches the performance of the specialized models on two out of three tasks and surpasses the no pretraining model finetuned on 100\% of the task-specific data. This highlights the suitability of contrastive pretraining for scenarios with limited task-specific data.
We analyse the impact of training on different layers in more detail in \cref{app:layers}, including the impact on individual task performance.

\subsection{Mixed-Text-Speech Input}\label{subsec:results-mixed-text-speech}

\begin{table}[t]
    \centering

    \resizebox{\linewidth}{!}{%
    \begin{tabular}{lccccc}
        \toprule
         \multirow{2}{*}{\textbf{Model}} & \multicolumn{1}{c}{\textbf{ASR}} & \multicolumn{1}{c}{\textbf{ST}} & \multicolumn{1}{c}{\textbf{SQA}}  & \textbf{Norm.} \\
        
              \cmidrule(lr){2-2} \cmidrule(lr){3-3} \cmidrule(lr){4-4} 
        
             & \textbf{WER$\downarrow$*} & \textbf{COMET$\uparrow$*} & \textbf{F1$\uparrow$*}  & \textbf{avg.$\uparrow$}\\
        \midrule

        no pretrain   & 23.78	&69.72	& 46.07	& -51.13\\
         ASR pretrain  & \textbf{12.21}	&75.70	&63.36	&  \hphantom{-}39.93 \\
        \midrule
        contr-cos-emb & 13.98	&76.32	&68.49	&  \hphantom{-}46.00\\
            \quad+ mixed  & 14.26	&76.61	&68.40	&   \hphantom{-}46.61\\
         
         contr-wasser-emb & 15.99	&77.09	&67.83	&  \hphantom{-}43.52 \\
        \quad+ mixed & 14.70	&75.05	&61.48 &   \hphantom{-}26.53 \\
        contr-cos-all & 13.06	& \textbf{78.29}	& \textbf{73.90}	&  \hphantom{-}\textbf{67.39}\\
          \quad+ mixed  &	17.63 & 73.83	& 55.12 &  \hphantom{-}\hphantom{0}2.14\\
        \midrule
        mixed-nwp & 15.27 & 	76.05	& 67.68	& \hphantom{-}39.67 \\   
        \bottomrule
    \end{tabular}%
    }
    \caption{Comparison of models pretrained on mixed speech-text input and finetuned on 10\% of the task-specific data.}
    \label{tab:nwp}
\end{table}

The previous section shows that contrastive learning is an effective pretraining strategy, outperforming ASR pretraining. We now analyse, whether aligning mixed speech-text and text sequences can further enhance performance.

\paragraph{Mixed Contrastive Pretraining.}
We adapt our approach to use mixed speech-text sequences (\textit{contr-*-*+mixed}). Using the \textit{contr-cos} loss on the embedding layer, our model matches the performance of standard contrastive pretraining, outperforming both no pretraining and ASR pretraining (\cref{tab:nwp}). However, with the \textit{contr-wasser} loss results degrade significantly. Consequently, we continue with \textit{contr-cos-all+mixed} to test the loss across all layers. This model also significantly underperforms its non-mixed counterpart. 

\begin{table}[t]
    \centering

    \resizebox{\linewidth}{!}{%
    \begin{tabular}{lccccc}
        \toprule
         \multirow{2}{*}{\textbf{Model}} & \multicolumn{1}{c}{\textbf{ASR}} & \multicolumn{1}{c}{\textbf{ST}} & \multicolumn{1}{c}{\textbf{SQA}}  & \textbf{Norm.} \\
        
              \cmidrule(lr){2-2} \cmidrule(lr){3-3} \cmidrule(lr){4-4} 
        
             & \textbf{WER$\downarrow$*} & \textbf{COMET$\uparrow$*} & \textbf{F1$\uparrow$*}  & \textbf{avg.$\uparrow$}\\
        \midrule

        ASR pretrain  & 12.21	&75.70	&63.36	&  \hphantom{-}39.93 \\
        \midrule
        contr-cos-emb & 13.98	&76.32	&68.49	&  \hphantom{-}46.00\\
        \quad+ asr loss & 13.18	& 76.89	& 68.67 & \hphantom{-}51.64 \\        
         contr-wasser-emb & 15.99	&77.09	&67.83	&  \hphantom{-}43.52 \\
         \quad+ asr loss & 13.19	& 77.40	& 70.44 & \hphantom{-}57.03\\
        \midrule
        contr-cos-all & 13.06	& 78.29	& 73.90 & \hphantom{-}67.39  \\
        \quad+ asr loss & 11.68	& 78.94	& 76.12	& \hphantom{-}78.17 \\
        contr-wasser-all & 12.92	& \textbf{80.52}	& \textbf{77.22} & \hphantom{-}84.96 \\
        \quad+ asr loss & \textbf{11.23}	& 80.26	& 77.19	 & \hphantom{-}\textbf{88.24} \\
        \bottomrule

    \end{tabular}%
    }
    \caption{Comparison of models pretrained on a combination of contrastive and ASR losses and then finetuned on a 10\% subset of task-specific data.}
    \label{tab:combined-loss}
\end{table}

\begin{table*}[ht]
    \centering
    \resizebox{\linewidth}{!}{%
    \begin{tabular}{clcccccccc}
        \toprule
        \textbf{FT} & \multirow{2}{*}{\textbf{Model}} & \multicolumn{1}{c}{\textbf{ASR}} & \multicolumn{2}{c}{\textbf{ST}} & \multicolumn{2}{c}{\textbf{SQA}}  & \multicolumn{2}{c}{\textbf{Contr loss on test}} & \textbf{Norm.} \\
        
        \cmidrule(lr){3-3} \cmidrule(lr){4-5} \cmidrule(lr){6-7} \cmidrule(lr){8-9}
        
       \textbf{Data} & & \textbf{WER$\downarrow$*} & \textbf{BLEU$\uparrow$} & \textbf{COMET$\uparrow$*} & \textbf{EM$\uparrow$} & \textbf{F1$\uparrow$*}  & \textbf{cos.} & \textbf{wasser.} & \textbf{avg.$\uparrow$}\\
        
        \midrule
        
        N/A & Specialized &  \hphantom{0}6.54 & 30.99 & 80.02 & 64.19 & 77.10 & N/A & N/A & \hphantom{-}100\\
        N/A  & BLSP-lslm-7b  &  44.51 &  28.70 &  78.68 & \hphantom{0}5.60	& 21.82 & N/A & N/A & \hphantom{0}-96.72\\
        N/A & Qwen2-Audio-7b & 12.03	& 21.57	& 74.81 & 27.79 &    46.75 &  N/A & N/A &  \hphantom{-}\hphantom{0}10.79 \\

         \midrule
        \multirow{6}{*}{10\%} & ASR pretrain  & 12.21	&24.82 &	75.70	& 49.48	& 63.36	& 1.37	& 1.30	&   \hphantom{0}\hphantom{-}39.93 \\
          &\quad+ giga & 12.16	& 27.44	& 78.38	& 62.23	& 73.17	& 1.37	& 1.31	&   \hphantom{-}\hphantom{0}69.37\\
        & contr-cos-all &  13.06 & 27.19	& 78.29	& 60.48	& 73.90	& 1.08	& 0.91	&  \hphantom{-}\hphantom{0}67.39\\
        &\quad+ giga & \textbf{10.94}	& \textbf{29.95}	& 81.21	& 65.45 & 79.12 & 1.06	& 0.71	&  \hphantom{-}\hphantom{0}97.12 \\
         & contr-cos-all + asr &   11.68	& 27.61	& 78.94	& 63.81	& 76.12	& 1.08	&0.91  &  \hphantom{-}\hphantom{0}78.17\\  
        &\quad+ giga & 11.12	& 29.90	& \textbf{81.29}	& \textbf{72.16}	& \textbf{82.52} & 0.98	& 0.65  & \hphantom{-}\textbf{102.15}\\
        \midrule
        \multirow{6}{*}{100\%} & ASR pretrain  & 10.28  &30.31&	80.98	&69.87	&80.92	&1.37	&1.22	&  \hphantom{-}100.24\\
         &\quad+ giga & 11.48	& 30.46	& 81.03	& 72.25	& 82.56	& 1.37	& 1.23	&   \hphantom{-}\hphantom{0}99.77 \\

        & contr-cos-all &  12.56	& 30.69	& 81.48	& 71.71	& 82.50	& 1.25	& 1.02	& \hphantom{-}\hphantom{0}99.06 \\
        &\quad+ giga &  \hphantom{0}\textbf{9.31}	& 31.36	& 81.98 &	75.12	& 84.67 &	1.10	& 0.83	& \hphantom{-}\textbf{114.18} \\
         & contr-cos-all + asr  & 10.04 &	30.82 &	81.56	& 72.40	& 82.47	& 1.19	& 1.03	& \hphantom{-}106.57\\
         &\quad+ giga & 10.03	& \textbf{31.54}	& \textbf{81.99}	& \textbf{76.11}	& \textbf{84.94}	& 1.10	& 0.79 & \hphantom{-}112.66 \\

        \bottomrule
    \end{tabular}%
    }
    \caption{Comparison of pretraining only on Must-C data or the combination with the bigger Giga dataset (+giga). We compare our models against BLSP \citep{wang2024blspbootstrappinglanguagespeechpretraining} and  the \texttt{Qwen2-Audio-7B-Instruct} model \citep{chu2024qwen2audiotechnicalreport}. Metrics with * contribute to the normalized average as in \cref{eq:norm_average}. Results for \textit{contr-wasser} can be found in \cref{app:more_giga} and do not outperform \textit{contr-cos}.\vspace{-0.2cm}}
    \label{tab:giga}
\end{table*}

\paragraph{Mixed NWP.}
We also propose using next-word prediction (NWP) with mixed speech-text inputs (\textit{mixed-nwp}) for pretraining, inspired by \citet{nguyen2024spiritlminterleavedspoken}. While this approach surpasses the no pretraining and ASR pretraining baselines (\cref{tab:nwp}, last row), it underperforms contrastive pretraining.
 
Further ablations of contrastive pretraining with mixed speech-text inputs and of mixed-nwp, including two different forced-alignment algorithms, do not yield additional improvements (detailed results, including all metrics, are provided in \cref{app:results_mixed_nwp}).

\subsection{Combining losses}

Building on the previous sections, which analyse contrastive pretraining and mixed speech-text pretraining in comparison to traditional ASR pretraining, we now explore the potential of combining these approaches. We sum the individual losses, assigning equal weight to each.

\paragraph{Contr. + ASR Loss.}
We find that combining the ASR pretraining loss with the contrastive losses on the embedding layer leads to improved performance for ASR and SQA, as seen in \cref{tab:combined-loss}. 
Similarly, this enhances the performance of the \textit{contr-*-all} models,
closing the gap between our contrastive pretrained models and ASR pretraining on ASR. 
\paragraph{Contr. + mixed-NWP Loss.}In contrast, combining the mixed-NWP loss with the contrastive losses does not yield further improvements. Detailed results are provided in \cref{app:combined_losses}.

\subsection{Additional pretraining data}\label{subsec:giga}
In the previous sections, we use Must-C ASR data for pretraining. Given the abundance of ASR data, we expand our pretraining dataset by incorporating GigaSpeech \citep{GigaSpeech2021}, effectively tripling the amount of data. The results demonstrate the significant impact of larger-scale pretraining.

\paragraph{Low-resource setting.}
As before, we use a 10\% finetuning subset to simulate a low-resource scenario. In preliminary experiments with lightweight contrastive pretraining on the embedding layer (\cref{app:more_giga}), we find that increasing the pretraining data leads to much larger improvements for \textit{contr-cos} loss ($+32.45$ Norm. avg.) than for \textit{contr-wasser} loss ($+6.60$). Therefore, we report \textit{contr-cos} results for subsequent experiments; results for \textit{contr-wasser} can be found in \cref{app:more_giga}. We perform contrastive learning on all layers with and without ASR loss, as these settings yielded the best results in the previous sections. As shown in \cref{tab:giga}, both models \textit{contr-cos-all+giga} and \textit{contr-cos-all+asr+giga} achieve an overall improvement of more than $19$ points (Norm. avg.) compared to the same models trained with less pretraining data, reaching $97.12$ and $102.15$ points respectively.
Remarkably, despite being finetuned on only 10\% of the task-specific data, these models surpass both specialized models, the Seamless model \citep{communication2023seamlessm4tmassivelymultilingual} for ST and the \citet{you-etal-2022-end} model for SQA. Moreover, they outperform the ASR-pretraining baseline even on the ASR task. 

\paragraph{Benchmarking our best variants.}
We evaluate our best-performing models in a higher resource setting.
We finetune \textit{contr-cos-all+giga} and \textit{contr-cos-all+asr+giga} on 100\% of the task-specific data. As expected, this leads to even stronger results. 
But while combining the contrastive and ASR losses provides significant improvements in the low-resource settings, when finetuning with 100\% of the available data, the 
contrastive loss alone performs on par with the combined loss. We hypothesize that the ASR loss is more sensitive to the out-of-domain Giga pretraining data than the contrastive loss, leading to no improvement on the ASR task.  

We compare these models against \texttt{blsp\_lsm\_7b} \citep{wang2024blspbootstrappinglanguagespeechpretraining}, which has a similar setup to ours in that it only trains the projector while keeping other components frozen, and \texttt{Qwen2-Audio-7b-Instruct} \citep{chu2024qwen2audiotechnicalreport}, a strong baseline SpeechLLM. Notably, BLSP is finetuned on Must-C translation data, whereas for all other tasks, BLSP and Qwen2-Audio results reflect zero-shot performance. Across all tasks, our model outperforms both of these SpeechLLMs. 

Results for \textit{contr-wasser} experiments can be found in \cref{app:more_giga}, these yield similar results but do not outperform \textit{contr-cos}.

\subsection{Impact of text-speech alignment on capturing paralinguistic features}
Speech data is much richer than a sequence of words, and the pretrained speech encoder (HuBERT) might capture additional paralinguistic features. 
By aligning the richer speech embeddings more closely with text embeddings, the pretraining methods we explore could potentially train the projector to discard these additional features. 

\begin{table*}[t]
    \centering
    \resizebox{\linewidth}{!}{%
    \begin{tabular}{llccccccc}
    \toprule

    \textbf{Model} & \textbf{Speaking} & \textbf{Gender} & \textbf{Noise} & \textbf{Reverberation} & \textbf{Speech} & \textbf{SDR} & \textbf{PESQ} \\
    & \textbf{Rate} & & & & \textbf{Monotony} & \textbf{Noise} & \textbf{Speech Quality} & \textbf{Avg.}\\
    \midrule

        random & 14.29 & 50.00 & 14.29 & 20.00 & 20.00 & 16.67 & 16.67 & 22.25 \\
        HuBERT + Class. Head & 57.38 & 64.69 & 27.89 & 28.12 & 39.35 & 42.85 & 27.22 & 41.07 \\
        \midrule
        no pretrain & \textbf{57.07}            &\textbf{64.26}      & \underline{27.62}   & 27.38         & \textbf{37.60}       & 33.86             & \underline{23.72}   & 38.79\\
        ASR pretrain & \textbf{57.23}         & \underline{57.95}   &\textbf{28.87}      & \textbf{29.05}          & 36.35                & \underline{30.19} & \underline{21.86} & \underline{37.36} \\
        contr-cos-all & 56.46       & 62.17              & 28.60                & \textbf{29.27}          & \underline{35.39}    & 34.17    & 24.12 & 38.60 \\
        contr-wasser-all & 56.12    & \textbf{62.30}     & 28.26               & 27.57        & \textbf{37.70}       & 34.25    & 24.30 & 38.64 \\
        contr-cos-all + asr & \underline{55.00} & 	61.29 &	\textbf{28.73} &	\underline{26.24} &	\underline{35.29} &	\underline{30.94} &	\textbf{24.86} &	\underline{37.48} \\
        contr-wasser-all + asr &  56.86 &	\underline{59.62} &	28.31 &	\textbf{29.21} &	36.64 &	\textbf{40.25} &	\textbf{26.37} &	\textbf{39.61} \\

    \bottomrule
    \end{tabular}%
    }
    \caption{Accuracy in \% results for different meta-speech categories on the mls-eng-speaker-descriptions dataset. Bold results are the best or not significantly worse than the best, underlined results are the weakest or not significantly worse  than the weakest result using pairwise t-test ($p<0.01$).}
    \label{tab:meta_results}
\end{table*}

\begin{table}[t]
    \centering
    \resizebox{\linewidth}{!}{%
    \begin{tabular}{lccc}
    \toprule
    \multirow{2}{*}{Model} & Avg. Gender & \multicolumn{2}{c}{COMET}  \\
    & Acc in \% & Correct & Diff \\
    \midrule
    HuBERT + LLama	& 72.21 & 77.24 & 0.69 \\
    no pretrain &  70.61 &	70.07	& 0.61 \\
    ASR pretrain	&74.11 &	77.46	& 0.84 \\
    contr-cos-all	& \textbf{78.42} &	\textbf{82.20}	& \textbf{1.09} \\
    contr-wasser-all	& 76.87 &	81.46	& 1.01 \\
    contr-cos-all + asr	& 76.10 &	80.01	& 0.95 \\ 
    contr-wasser-all + asr &  76.21 &	81.40	& 0.99 \\
    \bottomrule
    \end{tabular}%
    }
    \caption{Results for the MuST-SHE dataset \citep{bentivogli-etal-2020-gender} for different models. Following \citet{bentivogli-etal-2020-gender}, we evaluate the overall gender accuracy and the COMET score difference between the scores of correctly and incorrectly gendered references. Results are average over three language pairs, individual results can be found in \cref{tab:must_she_results}.}
    \label{tab:must_she_results_avg}
\end{table}

To assess whether this occurs, we examine our contrastively pretrained models on a) a paralinguistic classification task and b) a generation task where paralinugistic information is needed.

For a) we finetune our models on paralinguistic classification tasks before and after pretraining. In addition, we use the speech encoder (HuBERT) with a linear classification head as a baseline.
The results demonstrate that models with contrastive and ASR-based pretraining perform comparably to the baseline HuBERT model in predicting paralinguistic annotations, indicating that alignment-focused pretraining does not affect the retention of paralinguistic information significantly. Detailed accuracy results are presented in \cref{tab:meta_results}. In addition, \cref{tab:meta-mix} shows that incorporating these features as auxiliary finetuning tasks does not notably alter model performance.

For b) we test our models on the translation dataset MuST-SHE \citep{bentivogli-etal-2020-gender}, where the speakers's gender determines correct translations. We include a cascaded model (HuBERT + Llama) as a baseline where all paralinguistic features are lost due to the cascaded setup. We find that the gender accuracy of the translations is higher for contrastively pretrained models than for ASR pretrained models or the cascaded model as can be seen in \cref{tab:must_she_results_avg}, further supporting that contrastive pretraining does not hinder a model's ability to capture paralinguistic features.

\section{Conclusion}\label{sec:conclusion}
We introduce a contrastive learning-based pretraining strategy for SpeechLLMs, aligning text and speech embeddings in a parameter-efficient manner. This approach establishes an effective task-agnostic foundation for SpeechLLMs and demonstrates state-of-the-art performance across multiple downstream tasks, even when little finetuning data is available. 

Importantly, this pretraining method preserves the model's ability to understand paralinguistic information, offering a promising path forward for advancing SpeechLLMs. 

\section{Limitations}
While our work provides valuable insights into pretraining of LLMs, it has some limitations that offer opportunities for future research:

\paragraph{More Expensive Training Scenarios:} Our study focuses on a lightweight and flexible approach, keeping the LLM and speech encoder frozen while fine-tuning only the projector. Although this minimizes computational costs, future research could explore scenarios where the LLM and encoder or adapter layers are also fine-tuned, potentially unlocking additional capabilities.

\paragraph{Limited Incorporation of Meta-Speaker data:} We analyze whether contrastively pretrained models, designed to align speech and text, retain their ability to capture meta-speech information such as speaker characteristics or noise. While we find that these models do not lose this capacity in both classification and generation tasks, extending this analysis to additional generation tasks 
would provide a deeper understanding of their impact.

\paragraph{Low-resource Scenarios:}
Our study simulates low-resource settings by using subsets of the fine-tuning data, finding that models perform well even with only 10\% of the task-specific data. However, extending this analysis to truly low-resource languages or tasks would offer richer insights into the model's adaptability and performance under resource constraints.

\paragraph{Analysis of Contrastive Examples:} In this work,  we use all contrastive pairs within a minibatch to maintain an efficient implementation. However, an alternative approach could involve selecting contrastive examples that are more challenging by ensuring they are closer to the positive examples, as demonstrated by \citet{ye-etal-2022-cross}.

\paragraph{Potential Risks:} Enabling Language Models to understand speech makes them more easily and widely accessible. However, this also lowers the barrier for the potential misuse these models.  
In addition, SpeechLLMs can reproduce biases seen in the training data. We try to minimize this risk by not training the LLM, so any safeguard measures trained into the LLM also apply to the SpeechLLM. However, the model could still be affected by adversarial attacks.

\section*{Acknowledgments}
This work has received funding from the European Union’s Horizon research and innovation programme under grant agreement No 101135798, project Meetween (My Personal AI Mediator for Virtual MEETtings BetWEEN People). We gratefully acknowledge Poland’s high-performance Infrastructure PLGrid ACC Cyfronet AGH for providing computer facilities.

\bibliography{trimmed_anthology,custom.bib}
% \bibliographystyle{misc/acl_natbib}

%\bibliography{misc/custom}

\appendix
\section*{Appendix Overview}
The appendix includes the following information:
\begin{itemize}
    \item Hyperparameter and model details (\S\ref{app:hyper})
    \item Data details (\S\ref{app:data})
    \item Prompt details (\S\ref{app:prompts})
    \item Contrastive training over multiple layers (\S\ref{app:layers})
    \item Details on the mixed-nwp pretraining (\S\ref{app:mixed_nwp})
    \item Details on combining pretraining losses (\S\ref{app:combined_losses})
    \item Details on pretraining with more data (including translation results for all language pairs) (\S\ref{app:more_giga})
    \item Results for capturing paralinguistic features (\S\ref{app:meta-mix-results}) on classification as well as generation tasks

\end{itemize}

\section{Hyperparameters and Model Details}\label{app:hyper}
All hyperparameters, model and training parameters are listed in \cref{tab:hyperparameter}.

Our SpeechLLM architecture consists of three core components: a speech encoder, a text-based LLM, and a projector bridging the two. 
We select \texttt{facebook/hubert-large-ls960-ft} \citep{hubert-2021} as our speech encoder, \texttt{Llama-3.1-8B-Instruct} \citep{dubey2024llama3herdmodels} as our LLM and  Q-Former \citep{Li2023BLIP2BL} as the projector.

HuBERT has an \texttt{apache-2.0} license, the llama model has a \texttt{llama3.1} license. We adhere to both license requirements, when using the models for research.

For contrastive learning, we compare speech and text embeddings without the use of system prompts. This creates a positional mismatch between contrastive pretraining and fine-tuning. When using ROPE embeddings \citep{su2023roformerenhancedtransformerrotary}, this impacts contrastive pretraining at layers beyond the embedding layer. To address this potential overfitting problem, we experiment with varying the absolute starting position in RoPE by randomly sampling it during pretraining for each example.

\begin{table}[ht]
    \centering
    \resizebox{\linewidth}{!}{%
    \begin{tabular}{llc}
    \toprule
        training & Q-Former Num Query Token & 4 \\
        parameters & Q-Former Num Hidden Layers & 4\\
         & Q-Former Num Attention Heads & 12 \\
         & Q-Former Seconds per Window & 1/3 \\
         & per device batch size & 10 \\
         & gradient accumulation steps & 2 \\
         &  num GPUs & 4 \\
         & learning rate & 1e-4 \\
         & warmup ratio & 0.03 \\
         & optimizer & adamw\_torch \\
         & learning rate scheduler type & cosine \\
         & model max length & 2048 \\
         & gradient clipping & 1 \\
    \midrule
         pretraining  & num epochs & 5 \\
         specific& contrastive $\tau$ cos + wasser & 0.1 \\
         & contrastive $\tau$ nwp & 0.5 \\
         & sinkhorn loss p & 2 \\
         & sinkhorn loss blur & 0.5 \\
    \midrule
        finetuning  &  num epochs    & 2 \\
    \bottomrule
    \end{tabular}%
    }
    \caption{Hyperparameters for the trainings, which are conducted on four NVIDIA A100-SXM4-40GB GPUs, mostly following \citet{verdini2024connectspeechfoundationmodels}}
    \label{tab:hyperparameter}
\end{table}

\section{Data}\label{app:data}
We fine-tune and evaluate our contrastive pretraining method on three task: Automatic Speech Recognition (ASR), Speech Translation (ST), and Speech Question Answering (SQA). 

For ASR and ST, we use the MustC-v1 dataset \citep{di-gangi-etal-2019-must}. We use the en-de portion as English ASR data, and we use language pairs en-de, en-fr, en-it, and en-es for ST. We exclude examples longer than 45 seconds for training and use the \textit{tst-common} and \textit{tst-he} test sets for evaluation.

For question answering, we use Spoken-SQuAD \citep{lee2018spoken}, which is based on Stanford Question Answering Dataset \citep[SQuAD]{rajpurkar2016squad100000questionsmachine}. To allow for higher batch sizes during training, we exclude examples exceeding 150 seconds.

For pretraining, we use the MustC ASR data (approximately 400 hours), but also experiment with adding the M-version of Gigaspeech \citep{GigaSpeech2021}, which includes 1,000 hours of speech from audiobooks, podcasts, and YouTube.

Contrastive pretraining aligns text and speech, which could potentially lead to the loss of paralinguistic features in the embeddings. Therefore, we also train and  test our models on \textit{mls-eng-speaker-descriptions} \citep{Pratap2020MLSAL, lacombe-etal-2024-dataspeech, lyth2024natural}.
We use the \textit{mls-eng-speaker-descriptions} train and test set. The available training data is quite large, so we subsample it to make training feasible. Since the test set is predefined, we sample the training data such that the test data size is 5\% of the data. We let the model predict the following features: Speaking rate, gender, noise, reverberation, speech monotony, SDR noise,  and PESQ speech quality. Moreover, we test our models on MuST-SHE \citep{bentivogli-etal-2020-gender}, a translation test set, where the speaker gender is needed to translate the segment correctly. We use the en-es (), en-it and en-fr portion of MuST-SHE.

Details on prompts can be found in \cref{app:prompts}.

%\newpage
\section{Prompts}\label{app:prompts}
In this sections we report the prompt we used for the different tasks during inference. During training, for each example, the prompt is sampled from a list of multiple prompts. Here, we report the prompts used for testing, for the training prompts, we refer the reader to our codebase.

\begin{table}[ht]
    \centering
    \resizebox{\linewidth}{!}{%
    \begin{tabular}{ll}
    \toprule
      Task & Prompt \\
    \midrule
       ASR & Can you transcribe this audio? \\
       SQA & Listen to the audio and answer this question: \\
    \bottomrule        
    \end{tabular}%
    }
    \caption{ASR and SQA Prompts used for inference. We finetuned on a range of different prompts, please refer to our codebase for these.}
    \label{tab:prompts_ASR}
\end{table}

\begin{table}[ht]
    \centering
    \resizebox{\linewidth}{!}{%
    \begin{tabular}{p{0.1cm}p{10.5cm}}
    \toprule
     & Prompt \\
    \midrule
   en &  \textbf{Can you translate this audio from <sl> into <tl>?}\\

    fr &  Pouvez-vous traduire cet audio de <sl> en <tl>?\\
            it &      Puoi tradurre questo audio da <sl> in <tl>?\\

        es &  ¿Puedes traducir este audio de <sl> al <tl>?\\

       de    & Können Sie dieses Audio von <sl> zu <tl> übersetzen?\\

    \bottomrule        
    \end{tabular}%
    }
    \caption{ST Prompts used for inference. We finetuned on a range of different prompts, please refer to our codebase for these. For <sl> the source language is added, for <tl> the target language is added. }
    \label{tab:prompts_ST2}
\end{table}

\begin{table}[ht]
    \centering
    \resizebox{\linewidth}{!}{%
    \begin{tabular}{p{10cm}}
    \toprule
      Prompt \\
    \midrule
          Evaluate the audio file and provide details in the following order: speaking rate, gender, noise level, reverberation, speech monotony, SDR noise, and PESQ speech quality. Format the response as: <speaking\_rate>, <gender>, <noise>, <reverberation>, < speech\_monotony>, <sdr\_noise>, <pesq\_speech\_quality>. \\
      
    \bottomrule        
    \end{tabular}%
    }
    \caption{Prompts for the MLS speaker description dataset used for inference. We finetuned on a range of different prompts, please refer to our codebase for these.}
    \label{tab:prompts_meta}
\end{table}

\clearpage
\begin{figure*}[!ht]
    \centering
    \begin{subfigure}{0.49\textwidth}
        \centering
        \includegraphics[width=\linewidth]{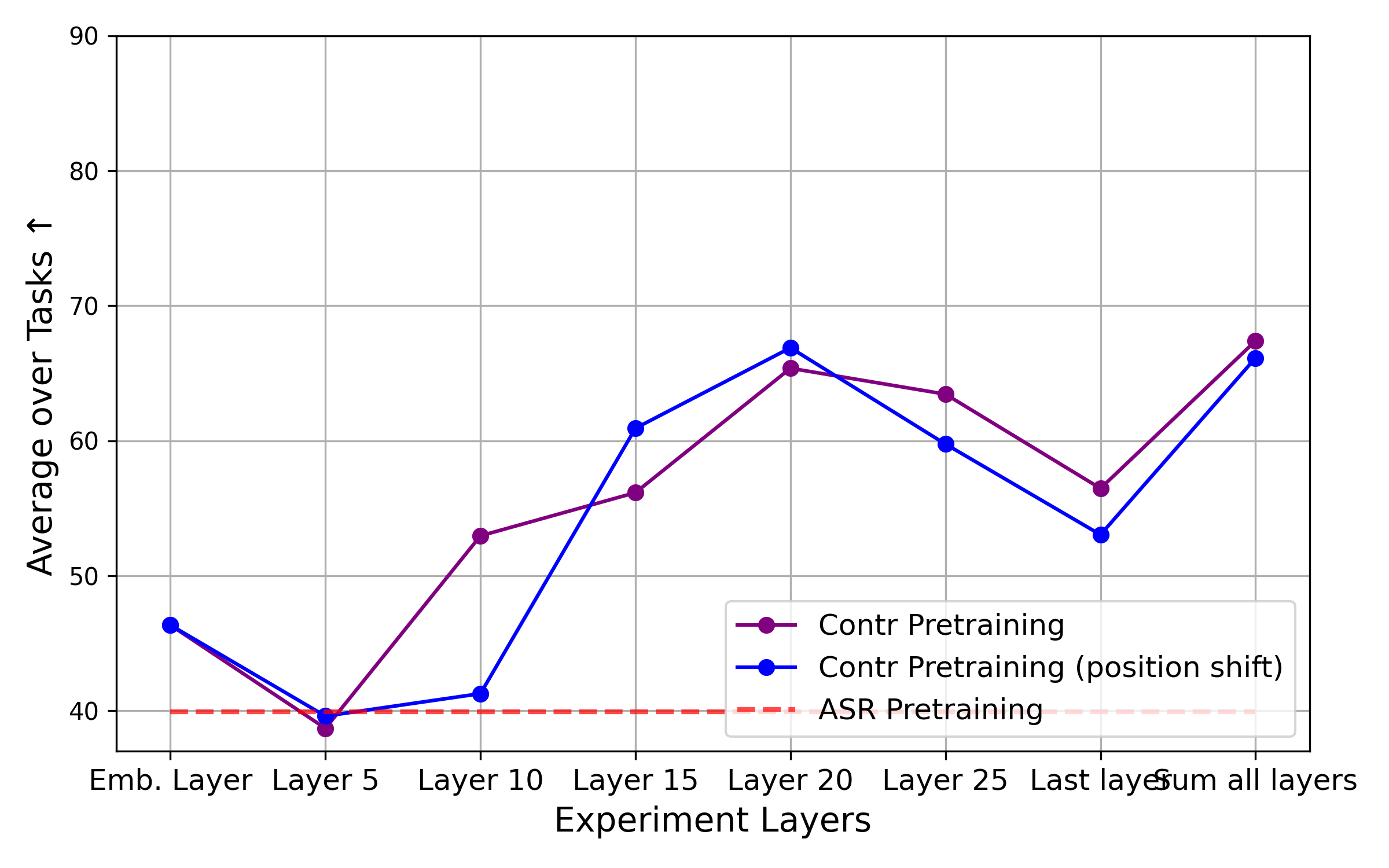}  % Replace with your image path
        \caption{contr-cos pretrained models}  % Caption for the first image  % Caption for the first image
        \label{fig:layers_avg}  % Label for referencing
    \end{subfigure}
    \hfill  % Optional spacing between the images
    \begin{subfigure}{0.49\textwidth}
        \centering
        \includegraphics[width=\linewidth]{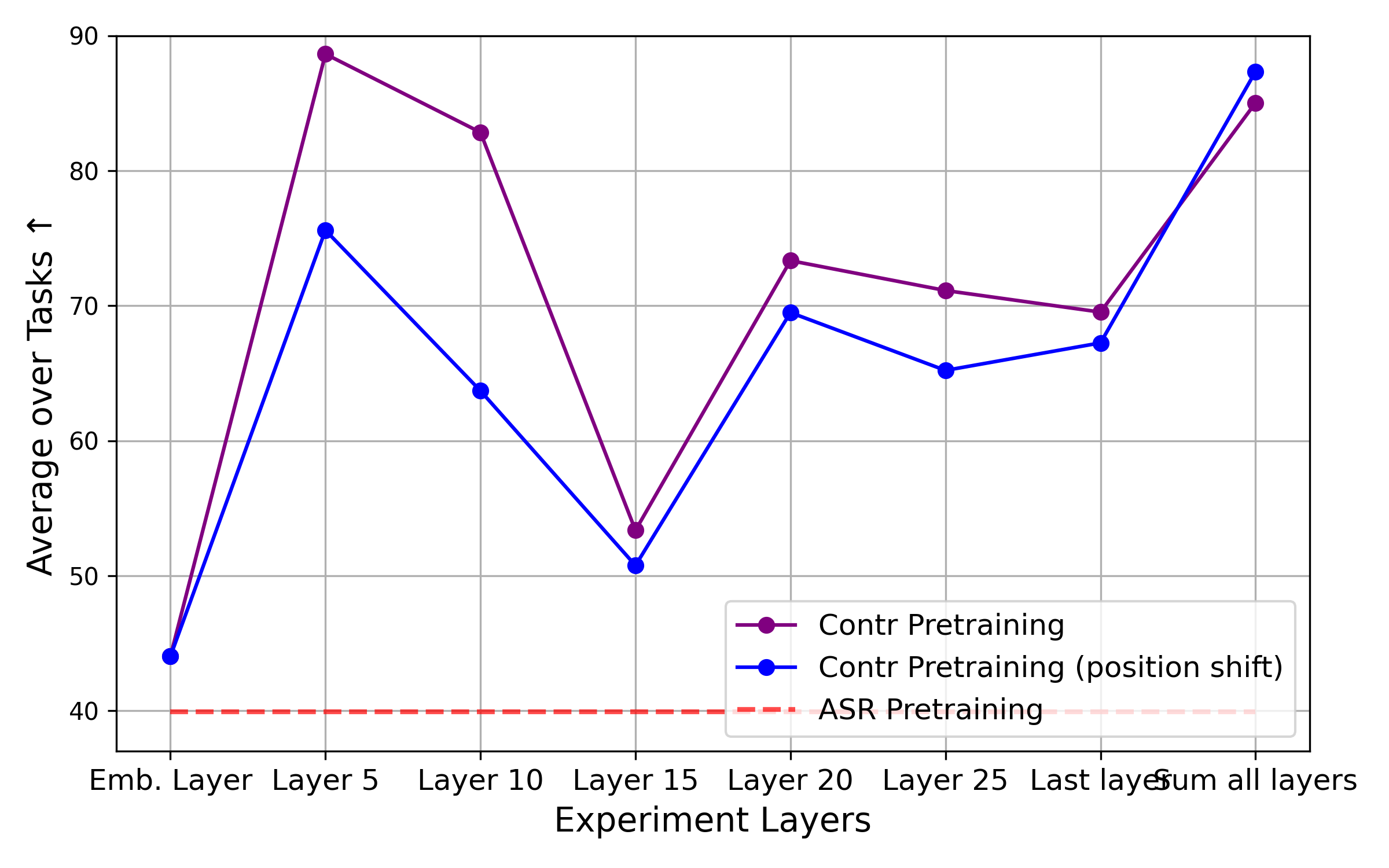}  % Replace with your image path
        \caption{contr-wasser pretrained models}  % Caption for the second image
        \label{fig:layers_wasser}  % Label for referencing
    \end{subfigure}
    \caption{Analysis of contrastive pretraining performed on different layers. We report the normalized average over tasks as in \cref{eq:norm_average} for contrastive pretraining and finetuning on 10\% of task-specific data. The last point on the x-axis refers to contrastive pretraining using the loss summed over all layers.}  
    \label{fig:contr_layers_all_tasks}
\end{figure*}
\begin{figure*}
    \centering
    \begin{subfigure}{0.49\textwidth}
        \centering
        \includegraphics[width=\linewidth]{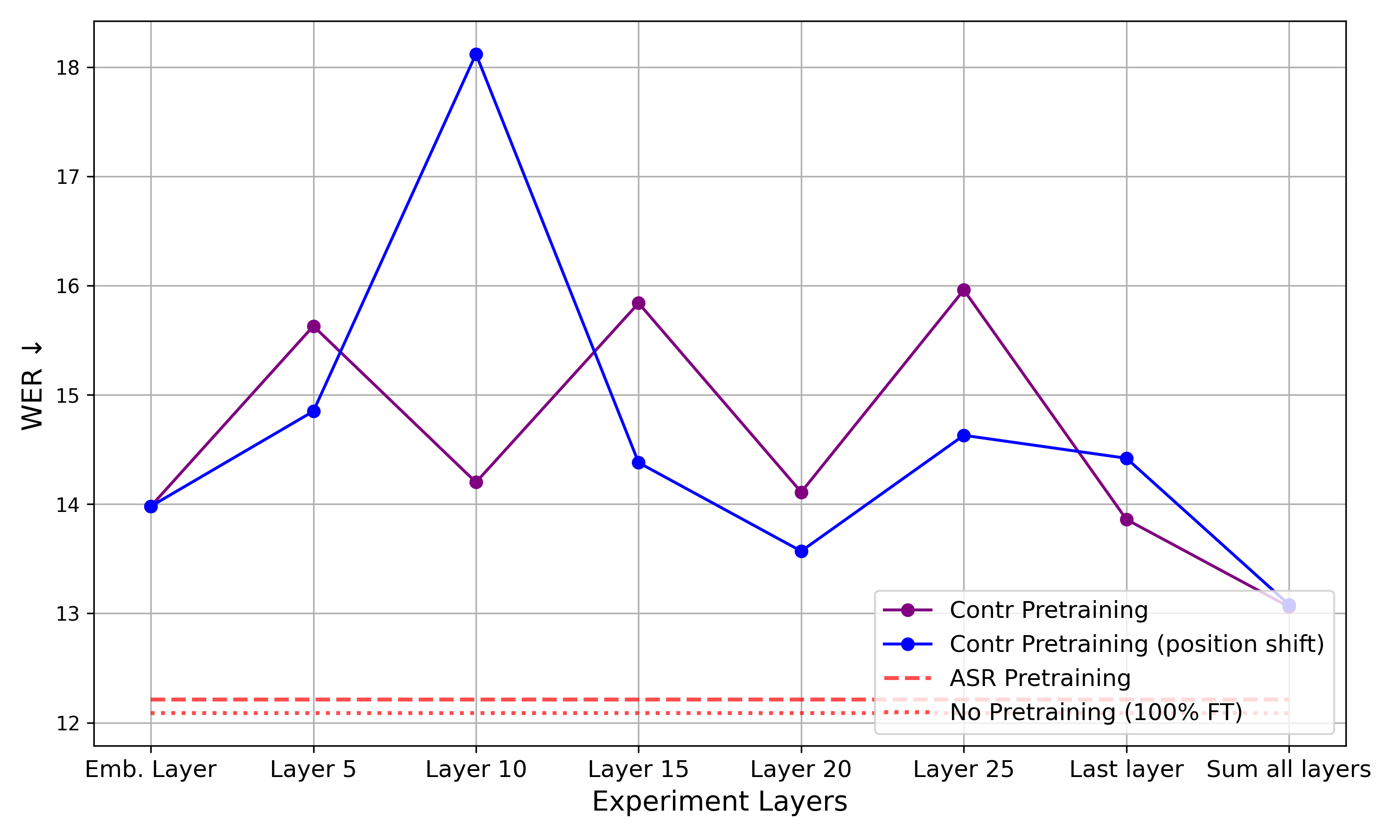}  % Replace with your image path
        \caption{ASR: contr-cos pretrained}  % Caption for the first image  % Caption for the first image
        \label{fig:loss_avg_asr}  % Label for referencing
    \end{subfigure}
    \hfill  % Optional spacing between the images
    \begin{subfigure}{0.49\textwidth}
        \centering
        \includegraphics[width=\linewidth]{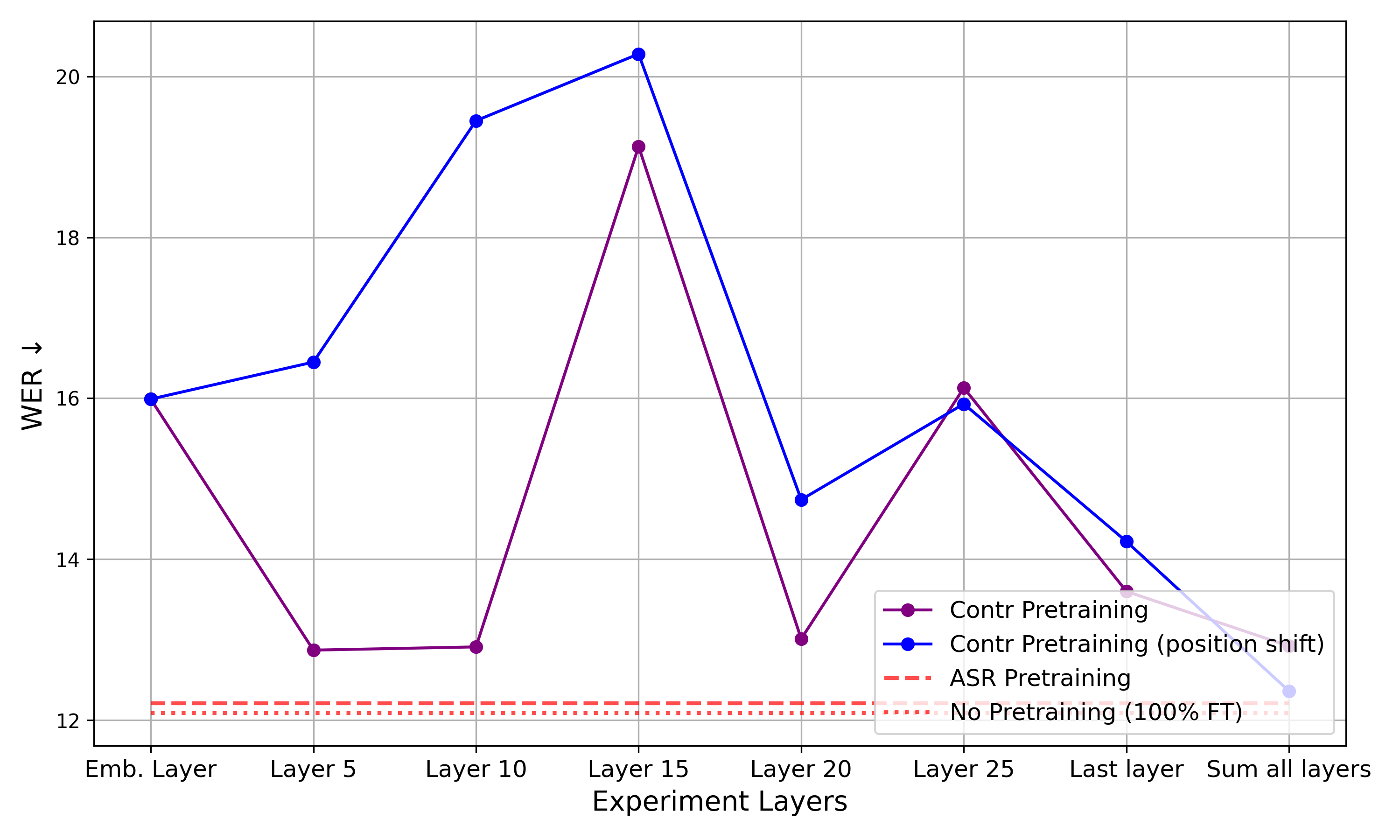}  % Replace with your image path
        \caption{ASR: contr-wasser pretained}  % Caption for the second image
        \label{fig:loss_wasser_asr}  % Label for referencing
    \end{subfigure}
    \hfill  % Optional spacing between the images
    \begin{subfigure}{0.49\textwidth}
        \centering
        \includegraphics[width=\linewidth]{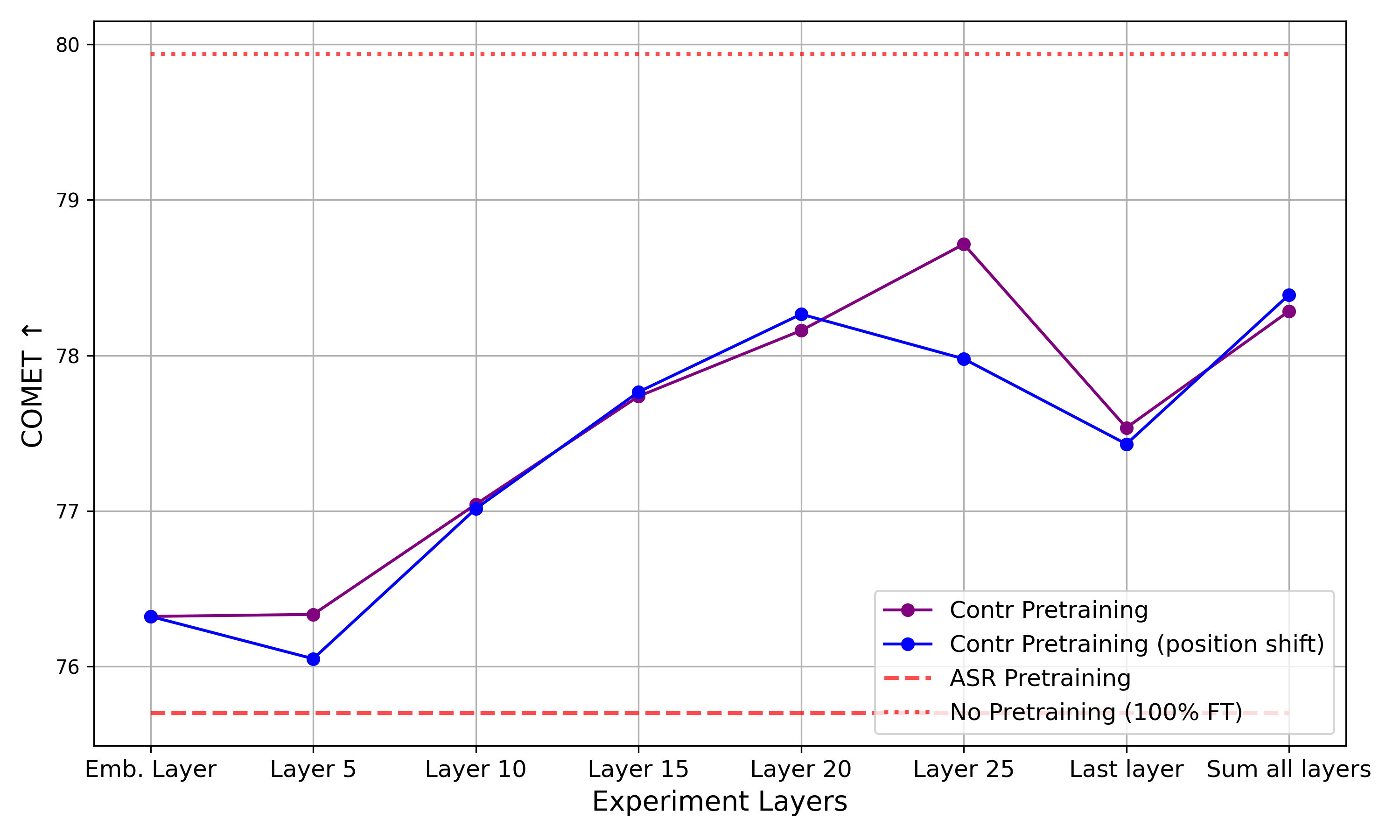}  % Replace with your image path
        \caption{ST: contr-cos pretrained}  % Caption for the first image  % Caption for the first image
        \label{fig:loss_avg_st}  % Label for referencing
    \end{subfigure}
    \hfill  % Optional spacing between the images
    \begin{subfigure}{0.49\textwidth}
        \centering
        \includegraphics[width=\linewidth]{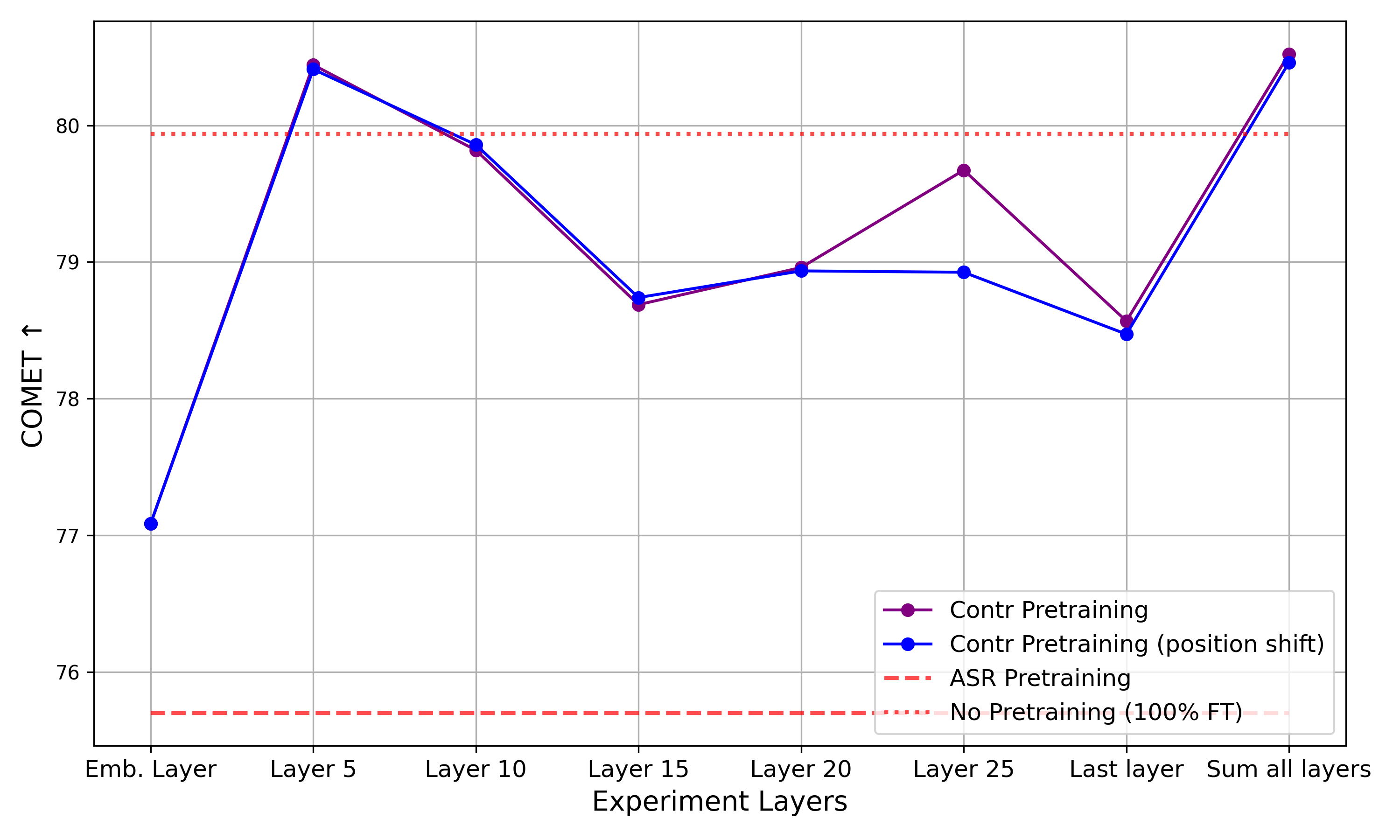}  % Replace with your image path
        \caption{ST: contr-wasser pretained}  % Caption for the second image
        \label{fig:loss_wasser_st}  % Label for referencing
    \end{subfigure}
    \hfill  % Optional spacing between the images
    \begin{subfigure}{0.49\textwidth}
        \centering
        \includegraphics[width=\linewidth]{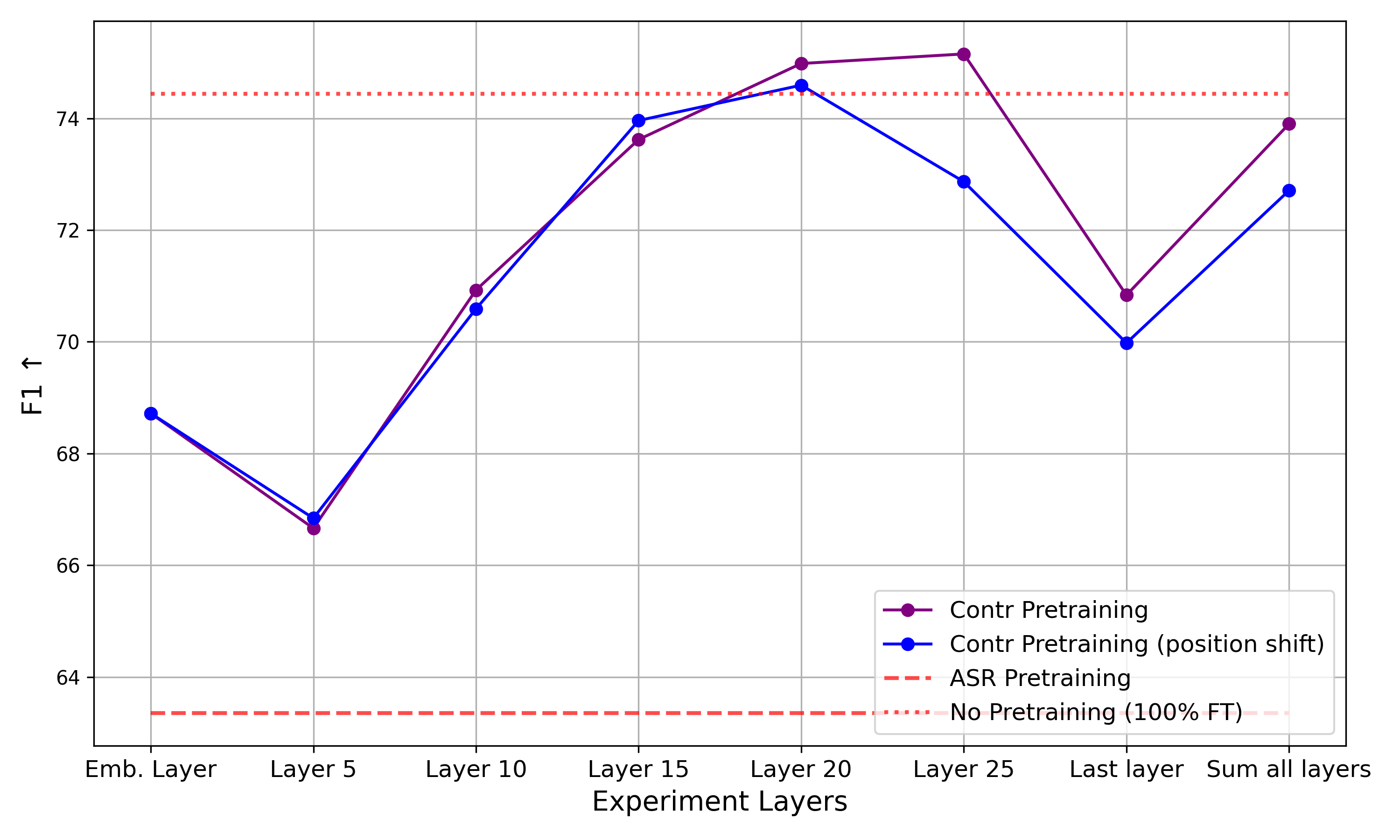}  % Replace with your image path
        \caption{SQA: contr-cos pretrained}  % Caption for the first image  % Caption for the first image
        \label{fig:loss_avg}  % Label for referencing
    \end{subfigure}
    \hfill  % Optional spacing between the images
    \begin{subfigure}{0.49\textwidth}
        \centering
        \includegraphics[width=\linewidth]{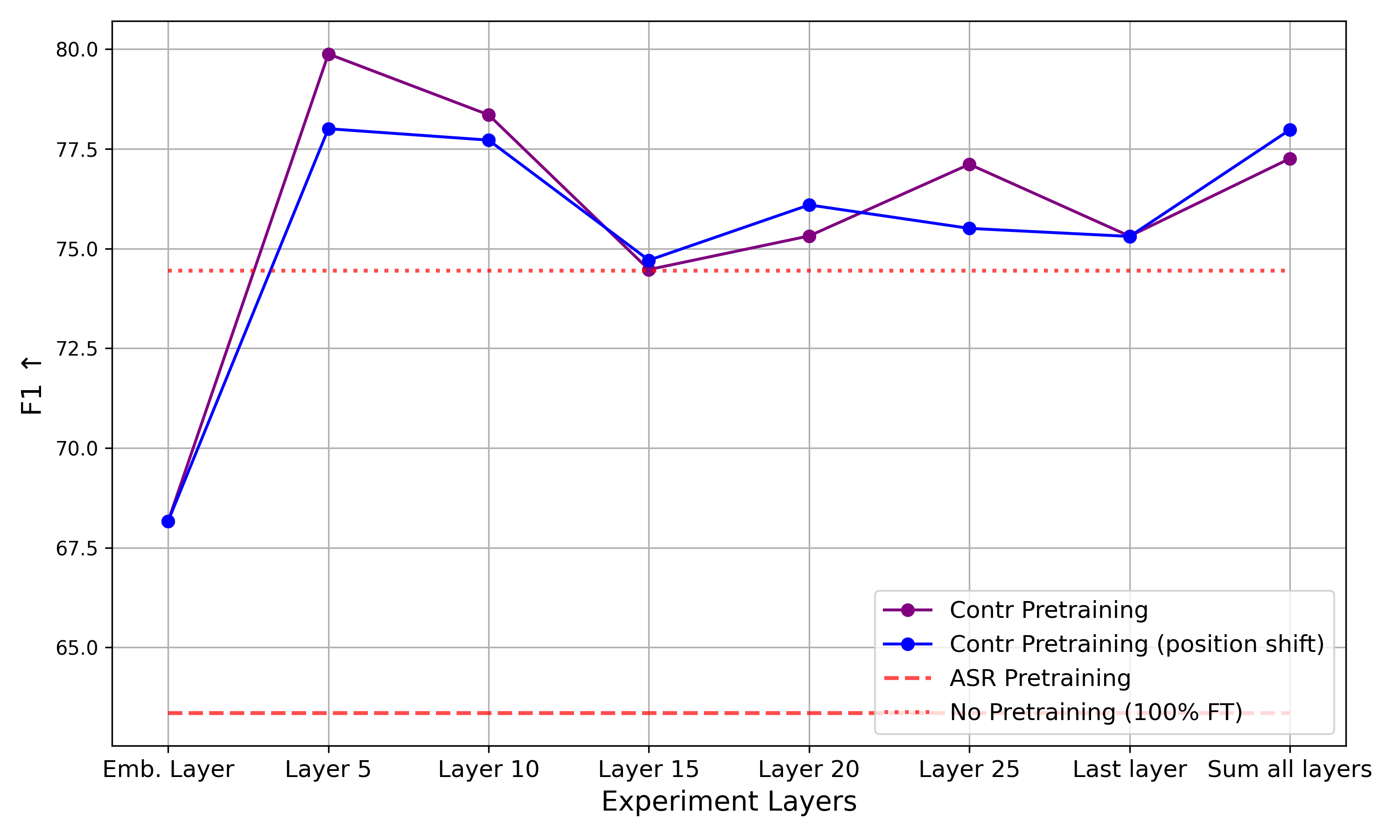}  % Replace with your image path
        \caption{SQA: contr-wasser pretained}  % Caption for the second image
        \label{fig:loss_wasser}  % Label for referencing
    \end{subfigure}
    \caption{Analysis of a contrastive pretraining performed on multiple layers for different tasks. The performance is reported after finetuning on 10\% of the task specific data. The last point on the x-axis refers to contrastive pretraining using the loss sum over all layers.}  
    \label{fig:contr_specific_layers}
\end{figure*}
\begin{figure*}
    \centering
    \begin{subfigure}{0.48\textwidth}
        \centering
        \includegraphics[width=\linewidth]{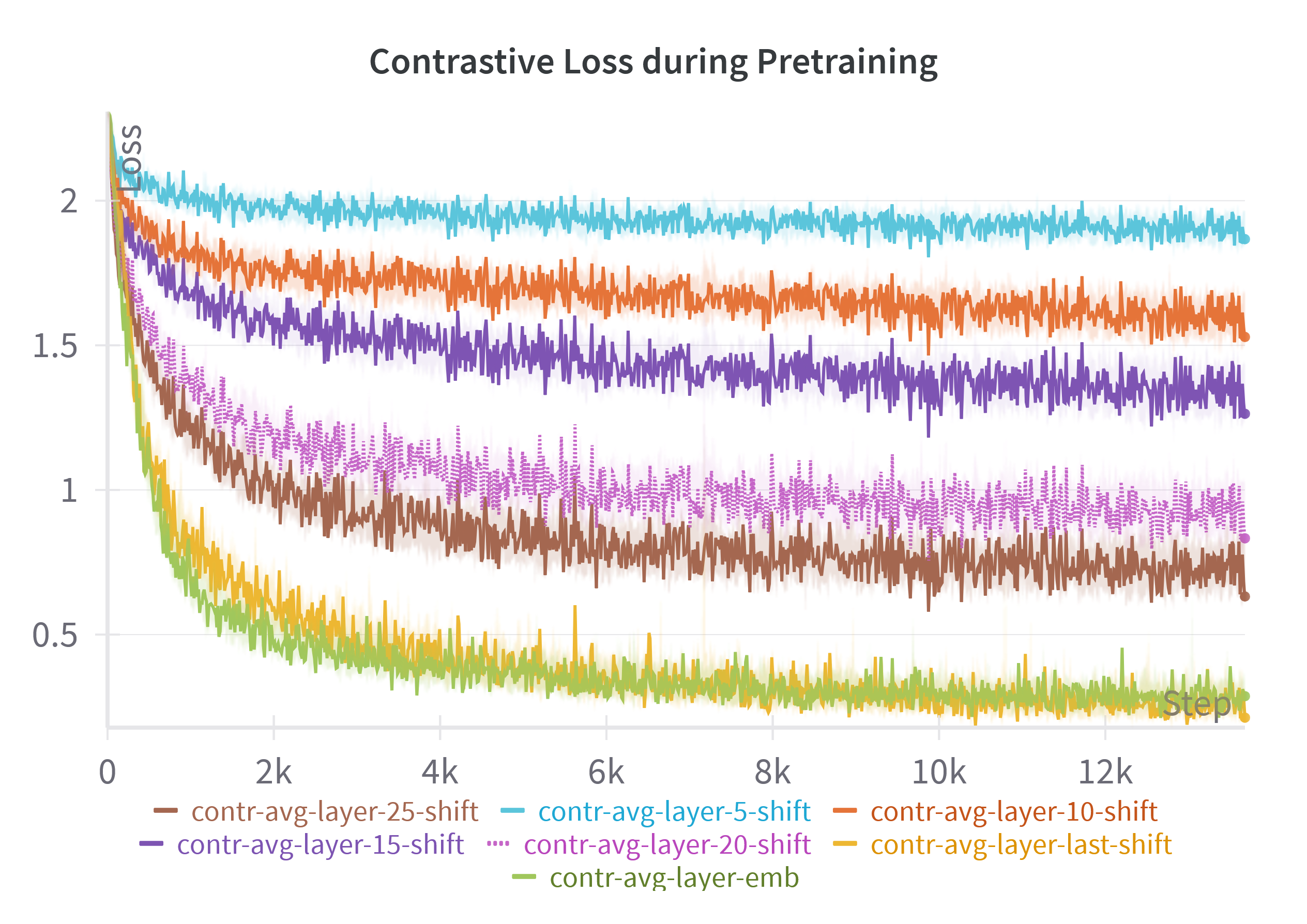}  % Replace with your image path
        \caption{Loss during contr-cos-petraining}  % Caption for the first image  % Caption for the first image
        \label{fig:layer_loss1}  % Label for referencing
    \end{subfigure}%
    \begin{subfigure}{0.48\textwidth}
        \centering
        \includegraphics[width=\linewidth]{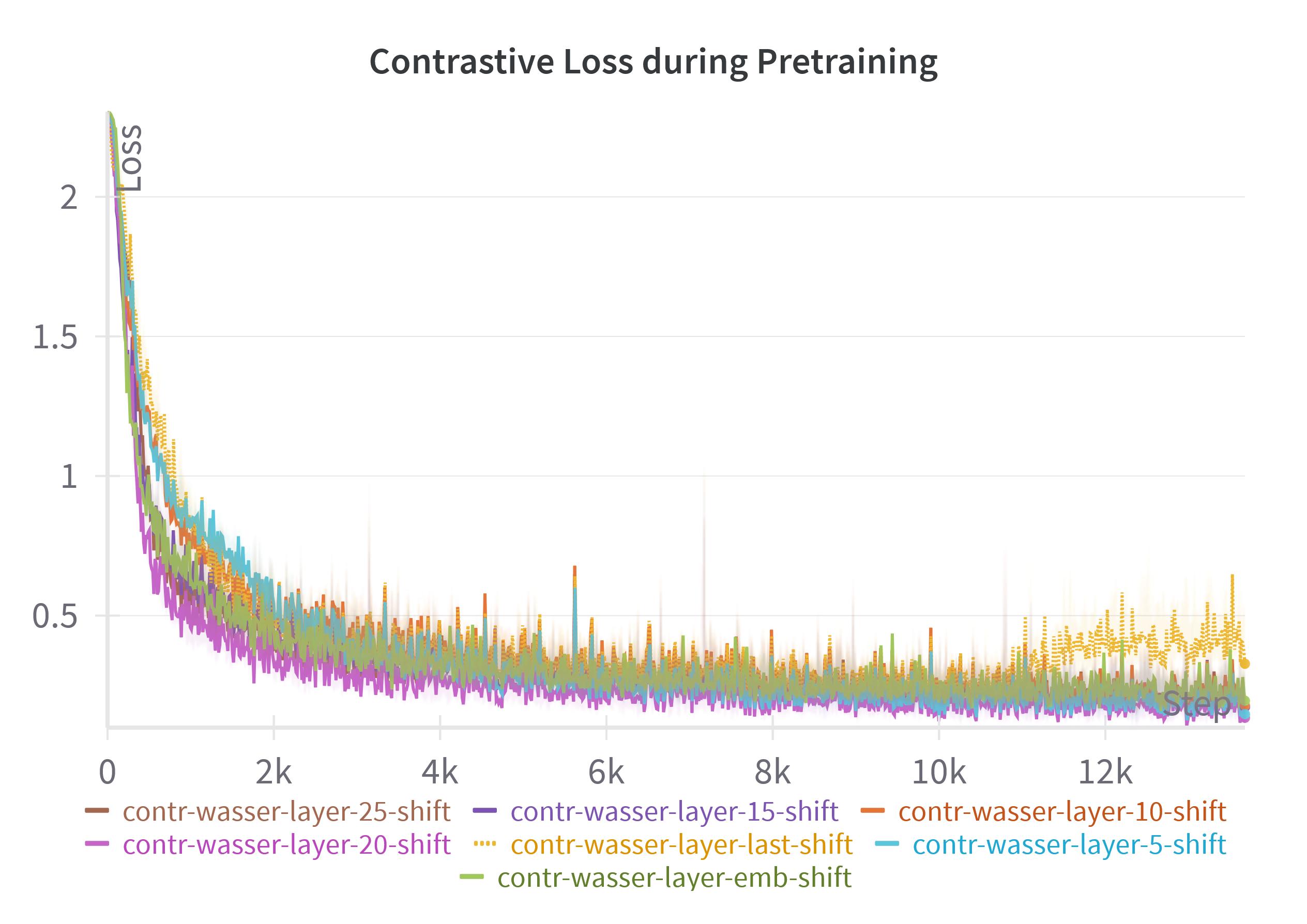}  % Replace with your image path
        \caption{Loss during contr-wasser-petraining}  % Caption for the second image
        \label{fig:layer_loss2}  % Label for referencing
    \end{subfigure}
    \caption{Contrastive loss during pretraining performed on multiple layers  for models trained with contrastive average loss (left) and contrastive wasserstein loss (right).}  
    \label{fig:contr_layers_loss}
\end{figure*}
\section{Contrastive Loss over Multiple Layers}\label{app:layers}
\begin{table*}[!ht]
    \centering

    \resizebox{\linewidth}{!}{%
    \begin{tabular}{clcccccccc}
        \toprule
        \textbf{FT} & \multirow{2}{*}{\textbf{Model}} & \multicolumn{1}{c}{\textbf{ASR}} & \multicolumn{2}{c}{\textbf{ST}} & \multicolumn{2}{c}{\textbf{SQA}}  & \multicolumn{2}{c}{\textbf{Contr loss on test}} & \textbf{Norm.} \\
        
        \cmidrule(lr){3-3} \cmidrule(lr){4-5} \cmidrule(lr){6-7} \cmidrule(lr){8-9}
        
       \textbf{Data} & & \textbf{WER$\downarrow$*} & \textbf{BLEU$\uparrow$} & \textbf{COMET$\uparrow$*} & \textbf{EM$\uparrow$} & \textbf{F1$\uparrow$*}  & \textbf{cos.} & \textbf{wasser.} & \textbf{avg.$\uparrow$}\\
        
        \midrule
        N/A & Specialized &   \hphantom{0}6.54 & 30.99 & 80.02 & 64.19 & 77.10 & N/A & N/A & 100\\
        N/A  & HuBERT + Llama & 18.38	  & 19.85	 & 73.92	& 36.15  &	54.76 & N/A &N/A &  0 \\
        100\% & no pretrain    &12.09 &	28.84	& 79.94	& 64.13	& 76.53	& 1.37	& 1.26	& 83.08 \\
        \midrule

        \multirow{2}{*}{10\%}  & contr-cos-all (multiple of 5) & \textbf{13.06}	&27.19	& 78.29	& 60.48	& 73.90	& 1.08 &	0.91	&  67.39\\
        & contr-cos-all  & 13.08	& \textbf{27.25}	& \textbf{78.39} & \textbf{61.45}	& \textbf{74.91} & 1.14	& 1.05& \textbf{69.41}\\
 
        \bottomrule
    \end{tabular}%
    }
    \caption{Initial ablations show that calculating the loss over multiples of 5 layers approximates calculating the loss over all layers efficiently.
    Results of models with contrastive pretraining using cosine similarity (\textit{contr-cos}) on all (\textit{-all}) layers, followed by fine-tuning on a 10\% subset of task-specific data.  The specialized baselines are  Whisper \citep{radford2022robustspeechrecognitionlargescale} for ASR, Seamless \citep{communication2023seamlessm4tmassivelymultilingual} for ST, and \citet{you-etal-2022-end} for SQA. Metrics with * contribute to the normalized average as in \cref{eq:norm_average}.}
    \label{tab:multiple_of_5_vs_all}
\end{table*}

\paragraph{Contrastive Loss over Different Layers}
For the contrastively pretrained model, we experiment with calculating the loss over different layers (\cref{fig:contr_layers_all_tasks}), using the final representation of each layer in the LLM.  We find that applying contrastive loss to deeper layers significantly increases downstream performance, however, the best layer to use is not consistent. Thus, using a general approach, such as calculating the loss over multiple layers, is the most likely to be robust to different settings. However, calculating the loss over all layers is computationally expensive. Specifically, when using the more computationally expensive Wasserstein distance, the total pretraining time increases by a factor of 4 when applied to all layers. In contrast, for the more efficient cosine similarity, the runtime only increases by 10\%. 
Initial experiments showed that using every 5th layer approximates the results well (\cref{tab:multiple_of_5_vs_all}), so we adopted this more efficient method for all experiments. 

\paragraph{Contrastive Loss over Different Layers on Different Tasks}
Further analysis reveals that contrastive learning on deeper layers benefit the ST and SQA  task the most, while improvements for ASR are less consistent.  \cref{fig:contr_specific_layers} shows the finetuning performance  on 10\% on task-specific data after pretraining on different layers for the tasks ASR, ST and SQA. 

\paragraph{Contrastive Loss Analysis during Pretraining}
\cref{fig:contr_layers_loss} shows the contrastive loss in the pretraining phase. The later the layer, the lower the contrastive loss, with an exception of the embedding layer. For cosine contrastive training, these differences are significantly bigger than for wasserstein contrastive training.
Here the losses for pretraining seem to correlated mostly with the performance of the model after finetuning: Models perform better after contrastive pretraining at later layers. However, for ASR, this trend is not so noticeable. 

\paragraph{Position Shifts during Pretraining}

\begin{table*}[ht]
    \centering

    \resizebox{\linewidth}{!}{%
    \begin{tabular}{clcccccc}
        \toprule
       \textbf{FT} & \multirow{2}{*}{\textbf{Model}} & \multicolumn{1}{c}{\textbf{ASR}} & \multicolumn{2}{c}{\textbf{ST}} & \multicolumn{2}{c}{\textbf{SQA}} & \textbf{Norm.} \\
        
        \cmidrule(lr){3-3} \cmidrule(lr){4-5} \cmidrule(lr){6-7} 
        
        \textbf{Data} & & \textbf{WER*} & \textbf{BLEU} & \textbf{COMET*} & \textbf{EM} & \textbf{F1*}  & \textbf{avg.}\\
        \midrule
        \multirow{2}{*}{10\%} & no pretrain   & 23.78	&19.87	&69.72	&34.72	&46.07	& -51.13\\
          &ASR pretrain  & \textbf{12.21}	&24.82 &	75.70	& 49.48	& 63.36	& \hphantom{-}39.93 \\
        \midrule
        \multirow{4}{*}{10\%} & mixed-nwp w/o punctuation (p) & \textbf{14.78}	& 23.85	& 75.78	& 47.35 & 66.39	& \hphantom{-}37.69 \\
         & mixed-nwp w/o punctuation (k) &  15.27 & 	25.22	& 76.05	& 53.57	& 67.68	& \hphantom{-}39.67 \\
        & mixed-nwp with punctuation (k) & 15.76	& 24.23	& 75.09	& 47.65	& 61.36		&  \hphantom{-}23.59  \\
        
        & mixed nwp w/o punctuation + giga (k) & 14.91	& \textbf{26.07}	& \textbf{77.17}	& \textbf{56.99}	& \textbf{70.94}	& \hphantom{-}\textbf{51.67}\\
        \midrule

         \multirow{5}{*}{10\%} &contr-cos-emb  + mixed (k) & \textbf{14.26}	&\textbf{25.96}	& \textbf{76.61}	&\textbf{51.97	}& \textbf{68.40}	&  \hphantom{-}\textbf{46.61}\\
        &contr-wasser-emb  + mixed (k) & 14.70	& 24.15	& 75.05	& 48.01	& 61.48 &  \hphantom{-}26.53 \\
         & contr-cos-last + mixed  (k) & 19.02	& 23.02	& 73.92	& 46.29	& 58.56	 & \hphantom{-}\hphantom{0}3.86 \\
        & contr-wasser-last + mixed (k) & 16.03	& 23.09	& 74.14	& 43.84	& 58.20	& \hphantom{-}12.95 \\
          & contr-cos-all + mixed  (k) &	
          17.63	& 22.90	& 73.83	& 42.15	& 55.12	&  \hphantom{-}\hphantom{0}2.14 \\

        \bottomrule
    \end{tabular}%
    }
    \caption{Ablation studies for models with mixed speech and text input. Metrics with * contribute to the normalized average as in \cref{eq:norm_average}. We experiment with two different forced alignment algorithms, one by \citet[][(p)]{pratap2023scalingspeechtechnology1000} and one by \citet[][(k)]{Kurzinger2020CTCSegmentationOL}. Since the latter performs better in our scenario, we use this aligner for all subsequent experiments.   }
    \label{tab:nwp_ablation}
\end{table*}

To address positional mismatches between pretraining and finetuning (detailed in \cref{app:hyper}), we experiment with position shifts during pretraining. Specifically, we experiment with varying the absolute starting position in RoPE by randomly sampling it during pretraining for each example. However, these do not consistently improve performance (\cref{fig:contr_layers_all_tasks}).

\section{Mixed Next Word Prediction}\label{app:mixed_nwp}
For alignment between text and speech, we utilize the forced alignment algorithm described by \citet{Kurzinger2020CTCSegmentationOL}, which leverages wav2vec 2.0 \citep{NEURIPS2020_92d1e1eb}. We also experimented with the alignment algorithm by \citet{pratap2023scalingspeechtechnology1000}, which performed slightly worse in our scenario (\cref{tab:nwp_ablation}).

We then randomly select spans from a sequence and assign them either text or speech embeddings, ensuring that sequences begin with text or speech equally often. Because speech embeddings are generally longer than text embeddings for the same content, we select spans of 2 to 5 words for speech and 4 to 10 words for text, inspired by \citet{nguyen2024spiritlminterleavedspoken} who use a similar distribution. However, unlike \citet{nguyen2024spiritlminterleavedspoken}, we do not introduce modality-specific tokens, as their framework uses these tokens to predict speech output — a capability we do not require.

In the alignment process for speech and text, the aligner generates timestamps for each word. If there is a pause between words, this pause is typically excluded from the alignment. However, to prevent the model from encountering unexpected silence or noise before a word during fine-tuning, we include any preceding silence or noise in the audio segment before the word.

\section{Mixed Speech-Text Inputs}\label{app:results_mixed_nwp}

We experiment with mixed speech-text input as described in \cref{subsec:mixed_nwp}, and report results in  \cref{tab:nwp_ablation}.

For the mixed speech-text input, we  experiment with two different forced alignment algorithms, one by \citet[][(p)]{pratap2023scalingspeechtechnology1000} and one by \citet[][(k)]{Kurzinger2020CTCSegmentationOL}. Since the latter performs better in our scenario, we use this aligner for all subsequent experiments.  

Pretraining with mixed speech-text next word prediction (nwp) achieves similar results as ASR pretraining. Here, removing punctuation seems to help the model understand the NWP task better.

We also test contrastive pretraining with mixed speech-text input (contr-*+mixed) and find that aligning the representation on the embedding layer works best (however, contrastive learning without mixed input still works better).

\begin{table*}[htbp]
    \centering

    \resizebox{\linewidth}{!}{%
    \begin{tabular}{llccccccc}
        \toprule
         \textbf{FT} & \multirow{2}{*}{\textbf{Model}} & \multicolumn{1}{c}{\textbf{ASR}} & \multicolumn{2}{c}{\textbf{ST}} & \multicolumn{2}{c}{\textbf{SQA}}  & \textbf{Norm.} \\
        
              \cmidrule(lr){3-3} \cmidrule(lr){4-5} \cmidrule(lr){6-7} 
        
            \textbf{Data}& & \textbf{WER$\downarrow$*} & \textbf{BLEU$\uparrow$} & \textbf{COMET$\uparrow$*} & \textbf{EM$\uparrow$} & \textbf{F1$\uparrow$*}  & \textbf{avg.$\uparrow$}\\
        \midrule
        10\% &  ASR pretrain  & 12.21	&24.82 &	75.70	& 49.48	& 63.36	& 39.93 \\       
        \midrule
           \multirow{6}{*}{10\%} &contr-cos-emb & 13.98	& 25.47	& 76.32	& 54.86	& 68.49	& 46.00\\
          & \quad+ asr loss & 13.18	& 25.77	& 76.89	& 56.12	& 68.67 & 51.64 \\        
          &\quad+ mixed + nwp & 14.68	& 25.28	& 76.18	& 48.86 & 63.92 & 36.41 \\
            &contr-wasser-emb & 15.99	& 26.10	& 77.09	& 54.64	& 67.83 &  43.52\\
          &\quad+ asr loss & 13.19	& 26.55	& 77.40	& 59.02	& 70.44 & 57.03\\
           & \quad+ mixed + nwp &  16.00	& 23.98	& 74.59	& 45.00	& 59.11 & 16.87\\

        \midrule
           \multirow{6}{*}{10\%} &contr-cos-all & 13.06	& 27.19	& 78.29	& 60.48	& 73.90 & 67.39  \\
         &\quad + asr loss & 11.68	& 27.61	& 78.94	& 64.16	& 76.12 & 78.17 \\
           &\quad+ mixed + nwp & 22.08	& 23.16	& 74.20	& 48.47	& 62.13	& \hphantom{0}2.07 \\
          & contr-wasser-all & 12.92	& \textbf{29.07}	& \textbf{80.52}	& 64.06	& \textbf{77.22} & 84.96 \\
         &\quad + asr loss & \textbf{11.23}	& 29.06	& 80.26	& \textbf{66.73}	& 77.19	 & \textbf{88.24} \\
           &\quad+ mixed + nwp & 15.72	& 24.13	& 75.00	& 51.21	& 64.25	& 	27.52\\
        \bottomrule
    \end{tabular}%
    }
    \caption{Comparison of models with combinations of alignment losses. Metrics with * contribute to the normalized average as in \cref{eq:norm_average}.}
    \label{tab:combined-loss_ablation}
\end{table*}

\section{Combined Losses}\label{app:combined_losses}

This section explores the combination of different losses. We combine 1) ASR and contrastive loss and 2) NWP and contrastive loss. 
The result can be found in \cref{tab:combined-loss_ablation}. While 1) leads to signifiant improvements, 2) does not seem to work together.

\section{Adding more pretraining data}\label{app:more_giga}
\paragraph{Results on 10\% of the data}
We analyize the effect of contrastive pretraining with more data, adding the Giga dataset to our pretraining data. Results for finetuning on 10\% of the data can be found in \cref{tab:giga_app}.
\begin{table*}[ht]
    \centering
    \resizebox{\linewidth}{!}{%
    \begin{tabular}{clcccccccc}
        \toprule
        \textbf{FT} & \multirow{2}{*}{\textbf{Model}} & \multicolumn{1}{c}{\textbf{ASR}} & \multicolumn{2}{c}{\textbf{ST}} & \multicolumn{2}{c}{\textbf{SQA}}  & \multicolumn{2}{c}{\textbf{Contr loss on test}} & \textbf{Norm.} \\
        
        \cmidrule(lr){3-3} \cmidrule(lr){4-5} \cmidrule(lr){6-7} \cmidrule(lr){8-9}
        
       \textbf{Data} & & \textbf{WER$\downarrow$*} & \textbf{BLEU$\uparrow$} & \textbf{COMET$\uparrow$*} & \textbf{EM$\uparrow$} & \textbf{F1$\uparrow$*}  & \textbf{cos.} & \textbf{wasser.} & \textbf{avg.$\uparrow$}\\
        
        \midrule
        N/A & Specialized &  \hphantom{0}6.54 & 30.99 & 80.02 & 64.19 & 77.10 & N/A & N/A & 100\\
        N/A  & BLSP-lslm-7b  &  44.51 &  28.70 &  78.68 & \hphantom{0}5.60	& 21.82 & N/A & N/A & -96.72\\
        N/A & Qwen2-Audio-7b & 12.03	& 21.57	& 74.81 & 27.79 &    46.75 &  N/A & N/A &  \hphantom{0}10.79 \\

        \midrule
        \multirow{14}{*}{10\%} & ASR pretrain  & 12.21	&24.82 &	75.70	& 49.48	& 63.36	& 1.37	& 1.30	&   \hphantom{0}39.93 \\
          &\quad+ giga & 12.16	& 27.44	& 78.38	& 62.23	& 73.17	& 1.37	& 1.31	&   \hphantom{0}69.37\\
         &  contr-cos-emb & 13.98	& 25.47	& 76.32	& 54.86	& 68.49	& 0.97	& 0.72	&  \hphantom{0}46.00\\
         &\quad+ giga & 14.24	& 28.76	& 80.18	& 63.94	& 76.60	& 0.97	& 0.47	&   \hphantom{0}78.45\\
         & contr-wasser-emb & 15.99	& 26.10	& 77.09	& 54.64	& 67.83	& 1.08	 & 0.62	&  \hphantom{0}43.52 \\
         &\quad+ giga & 16.36	& 26.47	& 77.59	& 58.05	& 71.09	& 1.06	& 0.58 &  \hphantom{0}50.12 \\
        & contr-cos-all &  13.06 & 27.19	& 78.29	& 60.48	& 73.90	& 1.08	& 0.91	&  \hphantom{0}67.39\\
        &\quad+ giga & \textbf{10.94}	& 29.95	& 81.21	& 65.45 & 79.12 & 1.06	& 0.71	&  \hphantom{0}97.12 \\
        & contr-wasser-all &  12.92	& 29.07	& 80.52	& 64.06	& 77.22	& 1.07& 	0.64	& \hphantom{0}84.96\\ 
        & \quad+ giga & 12.29	& 29.30	& 80.79	& 68.13	& 80.29	& 1.09	& 0.60	& \hphantom{0}92.75 \\
         & contr-cos-all + asr &   11.68	& 27.61	& 78.94	& 63.81	& 76.12	& 1.08	&0.91  &  \hphantom{0}78.17\\  
        &\quad+ giga & 11.12	& 29.90	& 81.29	& \textbf{72.16}	& \textbf{82.52} & 0.98	& 0.65  & \textbf{102.15}\\
        & contr-wasser-all + asr & 11.23	& 29.06	& 80.26	& 66.29	& 77.19	& 1.09 & 	0.66	& \hphantom{0}88.24 \\ 
        & \quad+ giga &  15.13	& \textbf{30.10} & 	\textbf{81.45} &	70.47 & 82.21 & 	1.03 &	0.30& 	\hphantom{0}91.26\\
        \midrule
        \multirow{10}{*}{100\%} & ASR pretrain  & 10.28  &30.31&	80.98	&69.87	&80.92	&1.37	&1.22	&  100.24\\
         &\quad+ giga & 11.48	& 30.46	& 81.03	& 72.25	& 82.56	& 1.37	& 1.23	&   \hphantom{0}99.77 \\

        & contr-cos-all &  12.56	& 30.69	& 81.48	& 71.71	& 82.50	& 1.25	& 1.02	& \hphantom{0}99.06 \\
        &\quad+ giga &  \hphantom{0}\textbf{9.31}	& 31.36	& 81.98 &	75.12	& 84.67 &	1.10	& 0.83	& \textbf{114.18} \\
        & contr-wasser-all  & 10.35	& 30.62 &	81.59 &	68.65& 	80.21& 	1.24 &	0.91 &	102.46\\
        & \quad+ giga & 12.10 & 	30.90	& 81.65 &	75.49	& 84.52	& 1.25	& 0.87&	104.32\\
         & contr-cos-all + asr  & 10.04 &	30.82 &	81.56	& 72.40	& 82.47	& 1.19	& 1.03	& 106.57\\
         &\quad+ giga & 10.03	& \textbf{31.54}	& 81.99	& \textbf{76.11}	& \textbf{84.94}	& 1.10	& 0.79 & 112.66 \\
        & contr-wasser-all + asr & 12.08	& 30.75 &	81.62	& 73.13 &	83.24 &	1.33	& 0.93 &	102.31\\
        & \quad+ giga & \hphantom{0}9.50	& 31.20 &	\textbf{82.00} &	74.27 & 	84.39 &	1.11	& 0.47 &	113.37 \\
        \bottomrule
    \end{tabular}%
    }
    \caption{Comparison of pretraining only on Must-C data or the combination with the bigger Giga dataset (+giga). We compare our models against BLSP \citep{wang2024blspbootstrappinglanguagespeechpretraining} and  the \texttt{Qwen2-Audio-7B-Instruct} model \citep{chu2024qwen2audiotechnicalreport}. Metrics with * contribute to the normalized average as in \cref{eq:norm_average}.}
    \label{tab:giga_app}
\end{table*}
\paragraph{Results on 100\% of the data}
The results on 100\% of the data can be found in \cref{tab:giga}

\paragraph{Results on different translation directions}

\begin{table*}[ht]
    \centering
    \resizebox{\linewidth}{!}{%
    \begin{tabular}{clcccccccccc}
        \toprule
        \textbf{FT} & \multirow{2}{*}{\textbf{Model}} & \multicolumn{4}{c}{\textbf{BLEU$\uparrow$} }  & \multicolumn{4}{c}{\textbf{COMET$\uparrow$*} }\\

       \textbf{Data} & &en-de &   en-es & en-fr & en-it & en-de &  en-es & en-fr & en-it \\
        
        \midrule
        N/A & Seamless & \textbf{26.25}	& \textbf{33.10}	& \textbf{36.49}	& \textbf{28.15}	& \textbf{79.03}	& \textbf{80.38}	& \textbf{79.92}	& \textbf{80.75}   \\
        N/A  & BLSP-lslm-7b  & 23.3\hphantom{0} & 27.4\hphantom{0} & 31.9\hphantom{0} & 23.2\hphantom{0} & 76.6\hphantom{0} & 78.7\hphantom{0} & 77.7\hphantom{0} & 78.2\hphantom{0}\\ 
        N/A & Qwen2-Audio-7b &19.60	 & 21.02	& 27.83	& 17.83	& 74.44 &	74.47	& 75.69	& 74.65 \\ 

         \midrule

        \multirow{6}{*}{10\%} & ASR pretrain  & 20.31	& 26.94	& 28.34	& 21.71	& 74.32	& 76.30	& 74.56 &	76.03  \\ 
          &\quad+ giga &23.25	& 29.81	& 32.42	& 24.28	& 77.56 &	79.12	& 77.68	& 79.17 \\

        & contr-cos-all & 23.23	& 29.20 & 	32.05	& 24.27	& 77.42	& 78.83	& 77.67	& 79.22 \\ 
        &\quad+ giga &25.29	& \textbf{32.68}	& \textbf{35.27}	& 26.55	& 80.44	& 81.64	& \textbf{80.97}	& 81.79  \\ 
         & contr-cos-all + asr &  23.51	& 29.85	& 32.62	& 24.46	& 78.25	& 79.42	& 78.43	& 79.66 \\ 
        &\quad+ giga &  \textbf{25.56}	& 32.34	& 35.01	& \textbf{26.67} & \textbf{80.63}	& \textbf{81.70}	& \textbf{80.97}	& \textbf{81.87} \\ 

        \midrule

        \multirow{6}{*}{100\%} & ASR pretrain  &	25.92	& 32.61	& 35.64	& 27.07	& 80.20	& 81.36	& 80.64	& 81.70\\
         &\quad+ giga &	26.05	& 32.55	& 35.90 & 27.33	& 80.40	& 81.25	& 80.61&	81.87  \\ % 81.03	

        & contr-cos-all & 	26.06	& 33.20	& 36.09	& 27.40	& 80.91	& 81.88 &	81.07	& 82.04 \\ %  81.48	
        &\quad+ giga &  	26.63	& \textbf{33.81}	& 36.90	& 28.20	& 81.23	& 82.30	& \textbf{81.74} & \textbf{82.63}	 \\ % 81.98
         & contr-cos-all + asr  & 	26.12 & 	33.23	& 35.98	& 27.93	& 80.86	& 82.02	& 81.24 &	82.11	\\ % 81.56
         &\quad+ giga & \textbf{26.88}	& 33.64	& \textbf{37.15}	& \textbf{28.49}	& \textbf{81.37}	& \textbf{82.34}	& 81.70	& 82.56  	\\ 

        \bottomrule
    \end{tabular}%
    }
    \caption{Comparison of translation performance of pretraining only on Must-C data or the combination with the bigger Giga dataset (+giga). We compare our models against BLSP \citep{wang2024blspbootstrappinglanguagespeechpretraining} and  the \texttt{Qwen2-Audio-7B-Instruct} model \citep{chu2024qwen2audiotechnicalreport}.}
    \label{tab:giga_translation_langs}
\end{table*}

For the models we discuss in \cref{subsec:giga}, we also report detailed results for the translation tasks. \cref{tab:giga_translation_langs} contains the results for the different translation directions.

\clearpage
\section{Impact of text-speech alignment of capturing paralinguistic features.}\label{app:meta-mix-results}

\vspace{-4.6cm}
Speech data is much richer than a sequence of words, and the pretrained speech encoder (HuBERT) might capture additional paralinguistic features. 
However, contrastive and ASR pretraining could potentially lead to discarding these paralinguistic features, when pulling speech embeddings closer to text embeddings.
To assess this, we test our models on a) on paralinguistic classification tasks and b) a generation task where paralinguistic information is needed.
\vspace{-1.5cm}
\paragraph{Classification Task on Paralinguistic Features}
For a) we finetune our models on paralinguistic classification tasks before and after pretraining. We also compare our models against a HuBERT + linear head baseline, to assess which features the encoder captures in the first place.
\vspace{-1.6cm}

To this end, we use the \textit{mls-eng-speaker-descriptions} train and test set \citep{Pratap2020MLSAL, lacombe-etal-2024-dataspeech, lyth2024natural}. This dataset contains classification tasks for rate, gender, background noise, and other features (more details in \cref{app:data}). 
The results demonstrate that models with contrastive and ASR-based pretraining perform comparably to the plain HuBERT model in predicting paralinguistic annotations, indicating that alignment-focused pretraining does not significantly affect the retention of paralinguistic information. \cref{tab:meta_results} shows the accuracies for the different classification tasks for the different models. 

\vspace{-1.5cm}
In a second step, we asses whether adding paralinguistic features during training (again using the mls-dataset), influences the models performance on downstream tasks. We finetune on a 10\% subset of  finetuning data to simulate a low-resource scenario. We find that including the paralinguistic data does not lead to significantly different results, which is not surprising given that the tasks that we test do not rely on paralinguistic information.  Results can be found in \cref{tab:meta-mix}.

\paragraph{Generation Task that Requires Paralinguistic Features}

For b) we test our models on MuST-SHE \citep{bentivogli-etal-2020-gender}, a translation dataset where the speaker's gender is necessary to correctly translate. We test the same models as for a)  but also include a cascaded model (HuBERT + Llama31Instruct) as a baseline where all paralinguistic features are lost due to the nature of the cascaded model.

The results can be found in \cref{tab:must_she_results}. Following \citet{bentivogli-etal-2020-gender}, we report the average gender accuracy using the evaluation script that comes with the data, and sensitivity, the score difference between the scores of correctly and incorrectly gendered references. We use the COMET difference (COMET Diff) here.

Across all language pairs, the contrastive models consistently outperform the cascaded model and the ASR pretrained model in terms of gender accuracy. On average across the language pairs, ASR pretraining performs 1.9\% points above the baseline, whereas the best performing contrastive model (cos-all) performs 6.2\% points above the cascaded baseline. The COMET Diffs are also larger for the contrastively pretained models than for ASR pretraining and the baseline. 

This underlines our findings that contrastive learning does not mitigate a model’s ability to reason about paralinguistic features. In contrast, ASR pretraining is not as effective at preserving this ability.
\begin{table*}[t]
    \centering

    %\resizebox{\linewidth}{!}{%
    \begin{tabular}{clcccccc}
        \toprule
        \textbf{FT} & \multirow{2}{*}{\textbf{Model}} & \multicolumn{1}{c}{\textbf{ASR}} & \multicolumn{2}{c}{\textbf{ST}} & \multicolumn{2}{c}{\textbf{SQA}}  &  \textbf{Norm.} \\
        
        \cmidrule(lr){3-3} \cmidrule(lr){4-5} \cmidrule(lr){6-7} 
        
       \textbf{Data} & & \textbf{WER$\downarrow$*} & \textbf{BLEU$\uparrow$} & \textbf{COMET$\uparrow$*} & \textbf{EM$\uparrow$} & \textbf{F1$\uparrow$*}  &  \textbf{avg.$\uparrow$}\\
        \midrule
        \multirow{2}{*}{10\%} & ASR pretrain  & 12.21	&24.82 &	\textbf{75.70}	& 49.48	& 63.36	& 39.93 \\
        &\quad + meta & \textbf{12.11}	& \textbf{24.91}	& \textbf{75.70}	& \textbf{49.47}	& \textbf{63.31} & \textbf{40.15}\\
        \midrule
        \multirow{4}{*}{10\%} & contr-cos-all &  13.06	&27.19	& 78.29	& 60.48	& 73.90&  67.39\\
        &\quad+ meta & 13.53	& 27.10	& 78.21	& 61.02	& 74.75 & 66.93\\        
         &contr-wasser-all & 12.92	& 29.07	& \textbf{80.52}	& 64.06	& 77.22 & 84.96\\
         &\quad+ meta & \textbf{11.97}	& \textbf{29.28}	& 80.48	& \textbf{66.12}	& \textbf{78.25}  & \textbf{88.92}\\
        \midrule
        \multirow{4}{*}{10\%} & contr-cos-all + asr &  11.68	& 27.61	& 78.94	& 63.81	& 76.12	&  78.17\\
        &\quad+ meta &  12.01	& 27.82	& 78.92	& 63.11	& 75.98 & 76.90\\
        &contr-wasser-all + asr & 11.23	& 29.06	& 80.26	& 66.29	&77.19	& 88.24 \\
        &\quad+ meta &  \textbf{11.16}	& \textbf{29.34}	& \textbf{80.55}	& \textbf{65.40}	& \textbf{78.26}	& \textbf{91.60}\\

        \bottomrule
    \end{tabular}%
    %}
    \caption{Comparison of models with and without pretraining, and with and without meta-speaker data in the finetuning process (on 10\%). Metrics with * contribute to the normalized average as in \cref{eq:norm_average}.}
    \label{tab:meta-mix}
\end{table*}

\begin{table*}[t]
    \centering
    \resizebox{\linewidth}{!}{%
    \begin{tabular}{llccc}
    \toprule
    Lang. Pair & Model & Avg. Gender Acc in \% & COMET correct & COMET diff \\
    \midrule
    \multirow{7}{*}{en-es}& HuBERT + LLama	& 73.52	& 78.69	& 0.7882 \\
    & no pretrain &  69.67	& 70.89	& 0.6510 \\
    & ASR pretrain	& 73.19	& 78.27	& 0.8523 \\
    & contr-cos-all	& 77.64	& \textbf{82.97}	& \textbf{1.1382 }\\
    & contr-wasser-all	& \textbf{76.73}	& 82.13& 	1.0784 \\
    & contr-cos-all + asr	& 76.14	& 80.67	& 1.0183 \\ 
    & contr-wasser-all + asr &  75.50	& 82.30	& 1.0809 \\
    \midrule
    \multirow{7}{*}{en-fr}& HuBERT + LLama	& 70.15	&75.13	&0.5278\\
    & no pretrain &   70.34	&68.66	&0.4591\\
    & ASR pretrain	& 72.64	&75.94	&0.6852 \\
    & contr-cos-all	& \textbf{77.83	}&\textbf{82.92}	&\textbf{1.1118}\\
    & contr-wasser-all	& 75.48	&79.97	&0.8856 \\
    & contr-cos-all + asr	& 75.73	&78.57	&0.8594\\ 
    & contr-wasser-all + asr  & 75.25	&79.96	&0.8909 \\
    \midrule
    \multirow{7}{*}{en-it}& HuBERT + LLama	& 72.96	&77.90	&0.7401 \\
    & no pretrain & 71.83	&70.66	&0.7281\\
    & ASR pretrain	& 76.50	&78.16	&0.9684  \\
    & contr-cos-all	&  \textbf{79.79}	&80.71	&1.0103 \\
    & contr-wasser-all	& 78.40	&\textbf{82.29}	&\textbf{1.0551}\\
    & contr-cos-all + asr	& 76.44	&80.78	&0.9682\\ 
    & contr-wasser-all + asr & 77.90	&81.93	&0.9858\\
    \bottomrule
    \end{tabular}%
    }
    \caption{Results for the MuST-SHE dataset \citep{bentivogli-etal-2020-gender} for different models. Following \citet{bentivogli-etal-2020-gender}, we evaluate the overall gender accuracy (using the evaluation script available with the dataset), and the score difference between the scores of correctly and incorrectly gendered references. We use the COMET difference (COMET Diff) here.}
    \label{tab:must_she_results}
\end{table*}

\end{document}